\documentclass[journal]{IEEEtran}

\usepackage{times}
\usepackage{amsmath,amssymb,amsfonts}
\usepackage{epsfig}
\usepackage{epstopdf}
\usepackage{graphicx}
\usepackage{subfigure}
\usepackage{color}
\usepackage{multirow,hhline}
\usepackage{url}
\usepackage{balance}
\usepackage{algorithm, algpseudocode, setspace}

\definecolor{darkgreen}{RGB}{0,160,0}
\newcommand{\x}     {{\bf x}}

\newcommand{\y}     {{\bf y}}
\newcommand{\RR}    {\mathbb{R}}
\newcommand{\ls}    {\hspace{2mm}}
\newcommand{\ms}    {\hspace{4mm}}

\newcommand{\given} {\, | \,}
\newcommand{\minus} {\! - \!}
\newcommand{\plus}  {\! + \!}
\newcommand{\ru}    {\rule{0mm}{4mm}}

\begin{document}

\title{A reliable order-statistics-based \\ approximate nearest neighbor search algorithm}
\author{Luisa Verdoliva, Davide Cozzolino, Giovanni Poggi
\thanks{Universit{\`a} Federico II di Napoli -- ITALY, e-mail {\{name.surname@unina.it\}}}
}
\maketitle

\markboth{IEEE Transactions on Image Processing}%
{Verdoliva \MakeLowercase{\textit{et al.}}: A reliable order-statistics-based approximate nearest neighbor search algorithm}

\begin{abstract}
We propose a new algorithm for fast approximate nearest neighbor search based on the properties of ordered vectors.
Data vectors are classified based on the index and sign of their largest components,
thereby partitioning the space in a number of cones centered in the origin.
The query is itself classified, and the search starts from the selected cone and proceeds to neighboring ones.
Overall,
the proposed algorithm corresponds to locality sensitive hashing in the space of directions,
with hashing based on the order of components.
Thanks to the statistical features emerging through ordering, it deals very well with the challenging case of unstructured data,
and is a valuable building block for more complex techniques dealing with structured data.
Experiments on both simulated and real-world data
prove the proposed algorithm to provide a state-of-the-art performance.
\end{abstract}


\section{Introduction}

A large number of applications in computer vision and image processing
need to retrieve, in a collection of vectors, the nearest neighbor (NN) to a given query.
A well-known example is image retrieval based on compact descriptors,
but there are countless more, from patch-based image denoising, to copy-move forgery detection, data compression, and so on.
Typically,
large sets of points are involved, calling for fast search techniques to guarantee an acceptable processing time.
However, for high-dimensional data, no {\em exact} NN search algorithm can provide a significant speed-up w.r.t. linear search,
so one is forced to settle for some {\em approximate} search algorithms,
trading off accuracy for efficiency.

In recent years, there has been intense research on techniques that improve this trade-off.
These can be classified in two large families according to their focus on memory or time efficiency.
A first family \cite{Jegou2011, Babenko2012, Gong2013, Ge2014a} addresses the case of very large datasets, that do not fit in memory.
This may occur, for example, in image retrieval and other computer vision applications.
In this condition, performing an accurate search based on Euclidean vector distances would require data transfers from disk, with an exceedingly large delay.
By associating compact codes with original vectors, and approximating exact distances based on such codes,
one can drastically reduce memory usage and avoid buffering,
with a huge impact on search speed.
In this case, therefore, memory efficiency is the main issue and the prevailing measure of performance,
while actual search time is of minor interest.

On the contrary, when data and associated structures can fit in memory,
processing time becomes the main issue, and performance is measured in terms of accuracy vs speed-up, with memory usage assuming minor importance.
This is the case of a large number of image processing application, especially in the patch-based paradigm.
Notable examples are nonlocal denoising \cite{Buades2005}, exemplar-based inpainting \cite{Guillemot2014},
copy-move forgery detection \cite{Cozzolino2015} and optical flow estimation \cite{Bao2014}.
The present work fits in the latter family.
That is, our aim is to provide a large speed-up with respect to linear search
while still guaranteeing a very high accuracy based on Euclidean distance computation on the original vectors.

Most techniques of this family follow a similar path, with many variations:
first the search space is partitioned into suitable cells\footnote{Often the cells overlap, so it is not a partition in strict sense:
this is an example of the many variations which we will overlook, from now on, for the sake of readability.},
and all vectors are classified based on the cell they belong to.
Then, at run time, the query is itself classified,
and the NN is searched only among vectors falling in the same cell as the query, or possibly in some neighboring ones.
Stated in these terms,
approximate NN (ANN) search bears striking similarities with the vector quantization (VQ) problem \cite{Gray1998,Gersho1992}.
In both fields,
defining a good partition of the space and a fast classification rule are the key ingredients of success.
Hence similar concepts and tools are used \cite{Pauleve2010}.

\begin{figure}

\centerline{
\includegraphics[width=2.2cm]{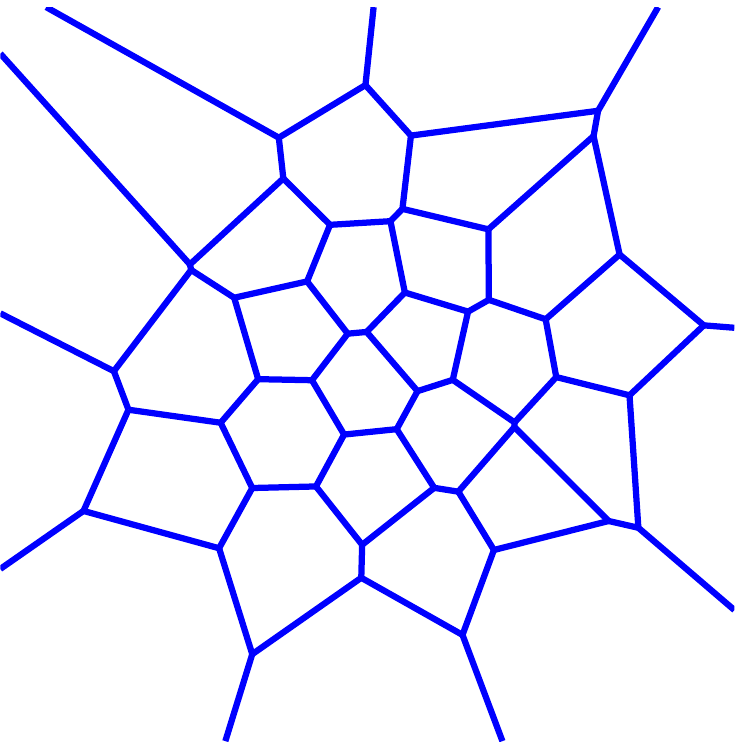} \ms
\includegraphics[width=2.2cm]{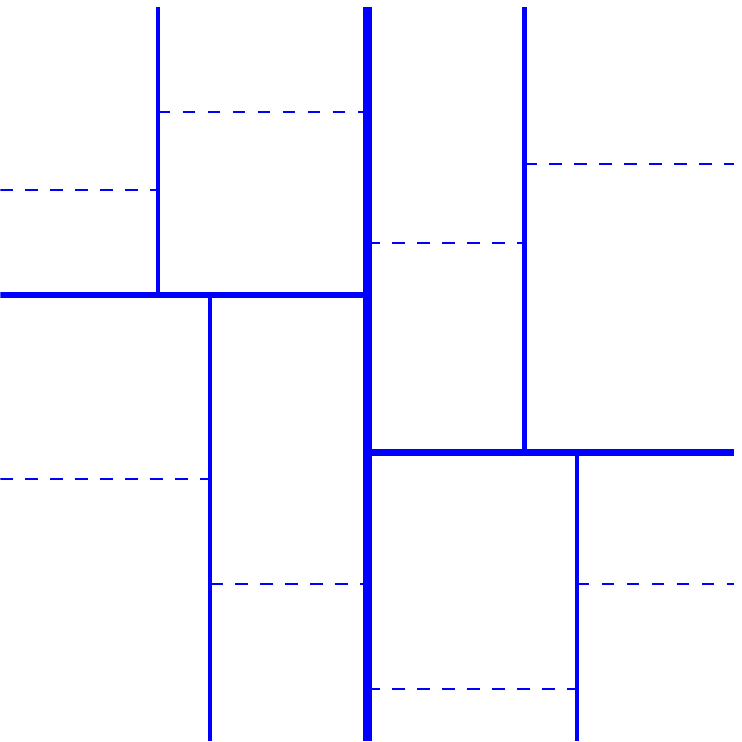} \ms
\includegraphics[width=2.2cm]{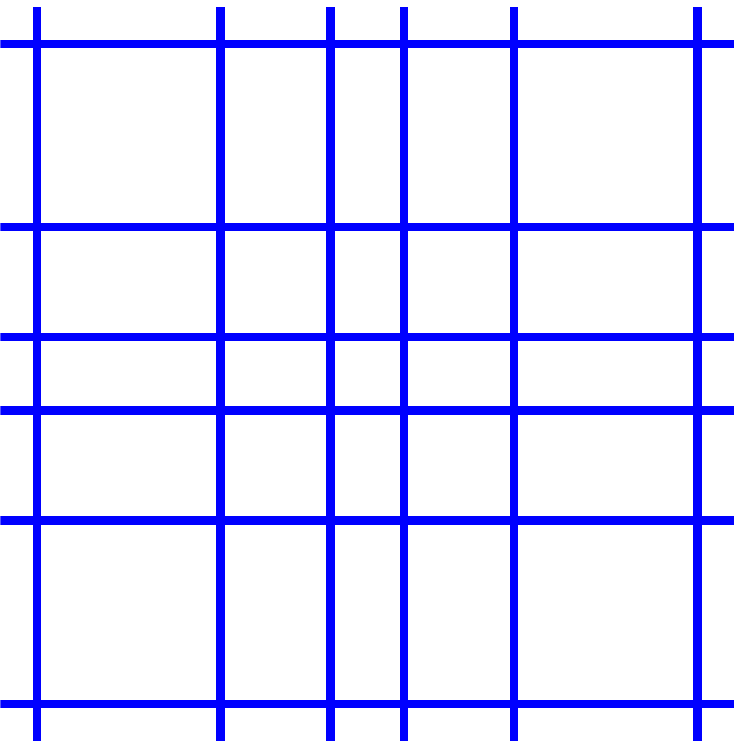} }

\vspace{1mm} \centerline{\small (a) \hspace{2.2cm} (b) \hspace{2.2cm} (c)}

\vspace{3mm} \centerline{
\includegraphics[width=2.2cm]{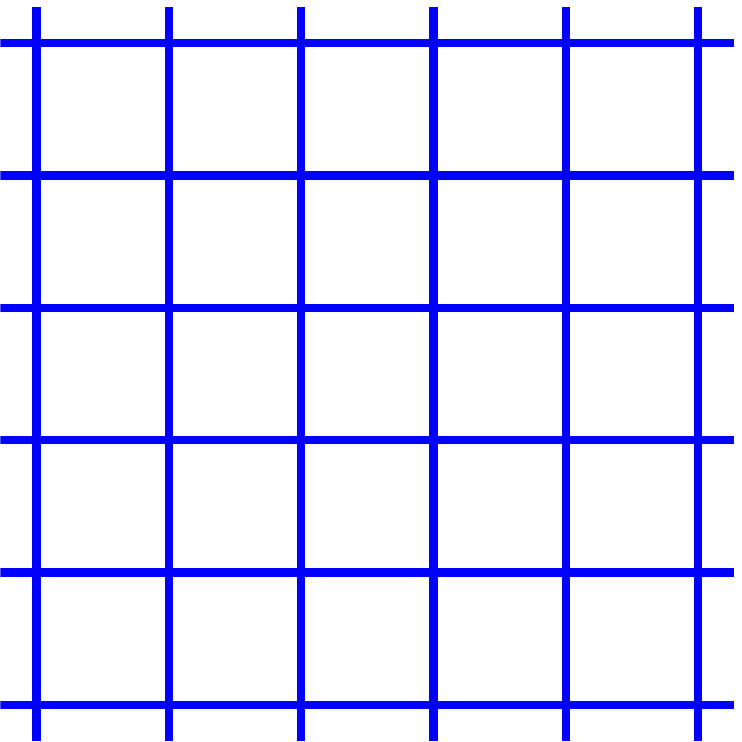} \ms
\includegraphics[width=2.2cm]{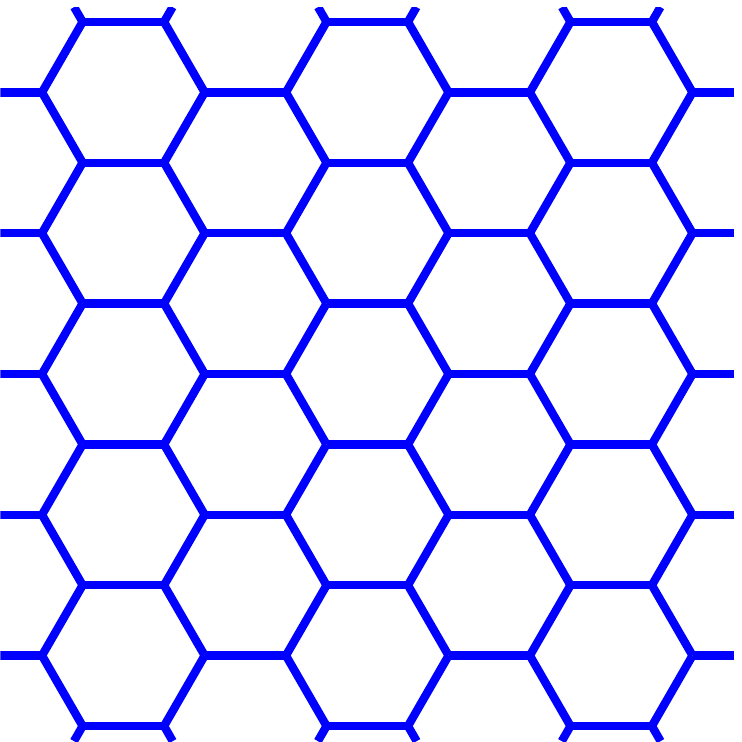} \ms
\includegraphics[width=2.2cm]{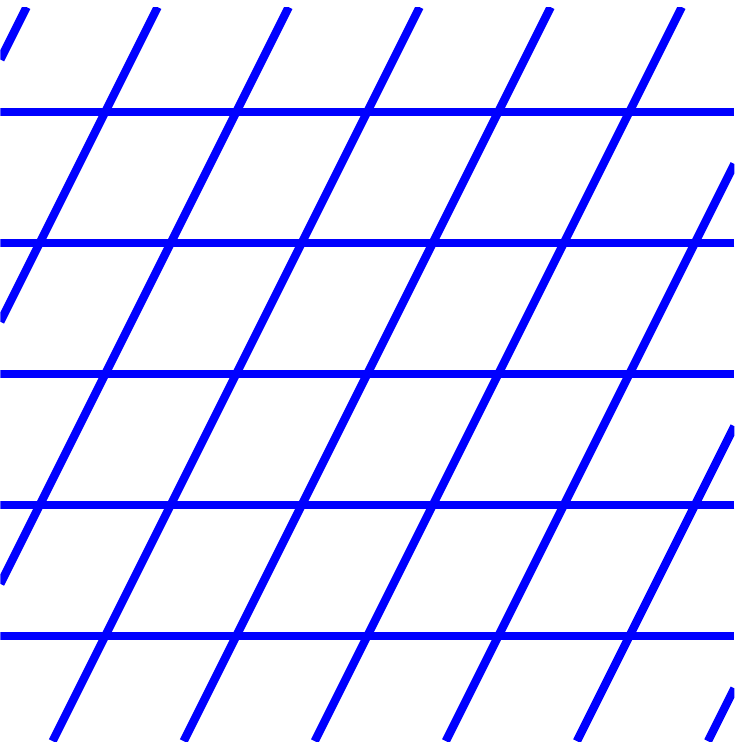} }

\vspace{1mm} \centerline{\small (d) \hspace{2.2cm} (e) \hspace{2.2cm} (f)}

\vspace{1mm}
\caption{ANN search is tightly related to quantization and space partitioning.
Some important concepts may emerge even with a simple 2d example.
For structured data, unconstrained vector quantization (a) is optimal but usually too complex for real-world applications.
Tree-structured VQ (b), and product VQ (c) reduce complexity but do not fully exploit data dependencies.
For unstructured data, it can make sense resorting to product scalar quantization (d).
However, lattice VQ (e) provides a better tessellation of the space.
Basic LSH corresponds to non-orthogonal product quantization (f), hence it is theoretically worse than both product  and Lattice VQ.
}
\label{fig:partitions}
\end{figure}

In the quantization literature there has been intense research on optimal (minimum distortion) space partitioning.
The k-means algorithm provides a locally optimal solution \cite{Lloyd1982}, with cells well packed in the space and adapted to data point density.
However,
to classify a query, $M$ vector distances must be computed, with $M$ the partition size, making this approach unsuited to fast NN search.
Therefore, a number of constrained solutions have been adopted in practice (see Fig.\ref{fig:partitions}).
In hierarchical k-means, a tree-structured search is carried out,
using partitions of size $M' \ll M$ at each node, with a sharp reduction of complexity.
The limiting case $M'=2$ corresponds to binary kd-trees search \cite{Bentley1975}.
With binary decisions at the nodes, the dataset is recursively partitioned by $k$-dimensional hyperplanes, allowing a very fast search.
However, complex tree-visiting schemes are needed to obtain an acceptable reliability.
In \cite{Anan2008, Muja2014} a good performance is obtained by using multiple randomized kd-trees.

Another popular approach is product quantization (PQ) \cite{Jegou2011},
with its many state-of-the-art variants,
inverted multi index \cite{Babenko2012},
Cartesian k-means \cite{Norouzi2013},
optimized PQ \cite{Ge2014a},
locally optimized PQ \cite{Kalantidis2014},
composite quantization \cite{Zhang2014}.
In PQ the vector space is decomposed into the Cartesian product of subspaces of lower dimensionality,
and quantization is carried out independently in each subspace.
Data points are then represented by compact codes obtained by stacking the quantization indices of all subspaces.
Thanks to these compact codes and to approximate code-based distances, large datasets can be dealt with efficiently.
The limiting case of scalar subspaces corresponds to the independent quantization of vector components.
This solution is simple but largely suboptimal,
and much better regular partitions of the space can be found \cite{Conway1998}.

%
%

Interestingly, one of the most popular ANN search techniques, locality sensitive hashing (LSH) \cite{Datar2004, Wang2014},
amounts to nothing more than scalar product quantization \cite{Pauleve2010}, with data projected on randomly oriented, non-orthogonal, axes.
Nonetheless, LSH can achieve a good performance through a number of clever expedients,
like the use of multiple hash tables \cite{Datar2004}, and multi-probe search \cite{Lv2007}.
Moreover, data-dependent variants of LSH improve largely over basic LSH,
by learning projection axes from the data \cite{Weiss2008, Brandt2010, Kong2012}, and using non-uniform quantizers \cite{Pauleve2010}.

In general,
taking advantage of the intrinsic structure of data can greatly help speeding up the search.
However, often data have little or no structure, or become unstructured after some preliminary processing.
In hierarchical k-means, for example, as well in IVFADC \cite{Jegou2011},
after the top-level quantization, the points become almost uniformly distributed in their cells.
Looking for the NN when data have no structure is an especially challenging task, encountered in many high-level problems,
and therefore a fundamental issue of ANN search.

In this work we propose a Reliable Order-Statistics based Approximate NN search Algorithm (ROSANNA) suited for unstructured data.
Like in the methods described before,
we define a suitable partition of the space, and carry out classification and priority search.
The main innovation consists in classifying the data based just on the index and sign of their largest components.
This simple pre-processing
allows us to partition the search space effectively, with negligible complexity and limited impact on memory usage.
For each query, only a short list of candidates is then selected for linear search.
ROSANNA can be also regarded as a variant of LSH, where the hashing is based on the vector directions.
This makes full sense in high dimensions, since data tend to distribute on a spherical shell.
ROSANNA produces a uniform partition of the space of directions,
and all vectors are automatically classified based on the order of their sorted components.
By using multiple hash tables, obtained through random rotations of the basis, and a suitable visiting scheme of the cells,
a very high accuracy is obtained, with significant speed-up w.r.t. reference techniques.
Experiments on both unstructured and real-world structured data prove ROSANNA to provide a state-of-the-art search performance.
We also used ROSANNA to initialize the NN field in copy-move forgery detection, a real-world computation-intensive image processing task,
obtaining a significant speed-up.


In the vast literature on ANN search, several papers related with ROSANNA have obviously appeared.
For example,
sorting has been already exploited by Chavez {\it et al.} \cite{Chavez2008}.
This technique, however, is pivot-based rather than partition-based.
A number of anchor vectors, or ``pivots'' are chosen in advance.
Then, for each database vector the distances to all pivots are computed and sorted,
keeping track of the order in a permutation vector.
At search time, a permutation vector is computed also for the query
and compared with those of the database points to select a short list of candidate NNs,
assuming that close vectors have similar permutation vectors.
In summary, sorting is used for very different goals than in the proposed method.

If ROSANNA is regarded as LSH in the space of directions,
then the same goal is pursued by the Spherical LSH (SLSH) proposed in \cite{Terasawa2007},
(not to be confused with the unrelated Spherical Hashing \cite{Heo2012}),
where a regular Voronoi partition of the unit hypersphere is built based on the vertices of an inscribed polytope.
For example,
working in a 3d space, the unit sphere can be partitioned in 8 cells corresponding to the 8 vertices of the inscribed cube.
Therefore, in SLSH, a regular partition is obtained with a low-complexity hashing rule \cite{Terasawa2007}.
Unfortunately,
in high-dimensional spaces ($K \geq5$) there are only three kinds of regular polytopes,
simplex, with $K$+1 vertices, orthoplex, with 2$K$ vertices, and hypercube with $2^K$ vertices.
SLSH uses eventually only the orthoplex polytope, which corresponds to a strongly constrained version of ROSANNA.
Likewise, Iterative quantization (ITQ), proposed in \cite{Gong2013} for ANN search through compact codes,
makes reference to hyper-octants, and hence corresponds to another (opposite) constrained version of ROSANNA.
By removing such constraints, ROSANNA is able to provide much better results.

In Concomitant LSH \cite{Eshghi2008}, instead,
a Voronoi partition of the space of directions is built based on a set of $M$ points taken at random on the unit sphere.
Although originally proposed for cosine distance,
it presents some similarities with ROSANNA, the use of directions and sorting, and can be easily adapted to deal with the Euclidean distance.
However,
to classify the query, $M$ distances must be computed at search-time, before inspecting the candidates,
which is a severe overhead for large hashing tables.
To reduce this burden, several variants are also proposed in \cite{Eshghi2008} which, however, tend to produce a worse partition of the space.
Irrespective of the implementation details,
the regular partition of the space of directions provided by ROSANNA can be expected to be more effective than the random partition used in Concomitant LSH.
Moreover, in ROSANNA, the hashing requires only a vector sorting, with no distance computation.
As for concomitants (related to order statistics) they are only used to carry out a theoretical analysis of performance,
but are not considered in the algorithm.

Recently, Cherian et al. proposed to use sparse ANN codes \cite{Cherian2014},
where each data point is represented as a sparse combination of unit-norm vectors drawn from a suitable dictionary.
The indexes of the selected dictionary vectors represent a short code used to speed up retrieval.
If a low-coherence dictionary is designed \cite{Cherian2014a},
close data points tend to fall in the same bucket, a cone identified by the selected dictionary vectors, which is then searched linearly.
This technique (SpANN) is explicitly designed to deal with datasets with large nominal and low intrinsic dimensionality (i.e., sparse).
Therefore, it is ineffective with unstructured data.
In this case, the extra efforts of designing a low-coherence dictionary (off-line),
and finding the sparse code of the query (at search time) are basically useless.
ROSANNA can be seen as the limiting case of SpANN when the vocabulary vectors are uniformly distributed over the space of directions
and a single vector is used to approximate a data point.

In the following,
we first describe the basic algorithm and explain its rationale (Section 2),
then describe the full-fledged implementation (Section 3),
discuss experiments on simulated and real-world data (Section 4)
and finally draw conclusions.

\section{ANN search based on order statistics}

\renewcommand{\ru}{\rule{0mm}{2.6mm}}
\newcommand{\bb}{\bf \color{blue}}
\newcommand{\bg}{\bf \color{darkgreen}}
\begin{figure}[t]
\centering

{\footnotesize
\begin{tabular}{|c|c||r|r|r|}\hline
\ru ~p~                  & c                  & $x_1$ & $x_2$ & $x_3$ \\ \hline\hline
\ru \multirow{7}{*}{1}   & \multirow{2}{*}{0} & {\bb  -22} &    12 &     5 \\
\ru                      &                    & {\bb  -21} &   -19 &   -12 \\ \cline{2-5}
\ru                      & \multirow{4}{*}{1} & {\bb   29} &    24 &   -13 \\
\ru                      &                    & {\bb   44} &    17 &    -4 \\
\ru                      &                    & {\bb   49} &    -6 &     5 \\
\ru                      &                    & {\bb   57} &     8 &    -2 \\ \hline\hline
\ru \multirow{6}{*}{2}   & \multirow{2}{*}{0} &    -3 & {\bb  -18} &    10 \\
\ru                      &                    &    -1 & {\bb  -13} &     0 \\ \cline{2-5}
\ru                      & \multirow{3}{*}{1} &     5 & {\bb   11} &     4 \\
\ru                      &                    &    11 & {\bb   14} &    -3 \\
\ru                      &                    &    14 & {\bb   25} &    23 \\ \hline\hline
\ru \multirow{6}{*}{3}   & \multirow{4}{*}{0} &   -36 &    23 & {\bb  -47} \\
\ru                      &                    &     5 &    26 & {\bb  -27} \\
\ru                      &                    &     9 &    -2 & {\bb  -17} \\
\ru                      &                    &    12 &     5 & {\bb  -14} \\ \cline{2-5}
\ru                      & \multirow{1}{*}{1} &    -7 &    11 & {\bb   22} \\ \hline
\end{tabular}
\;\;\;\;\;\;
\begin{tabular}{|c|c||r|r|r|}\hline
\ru p                    & c                  & $x_1$ & $x_2$ & $x_3$ \\ \hline\hline
\ru \multirow{9}{*}{1-2} & \multirow{2}{*}{0} & {\bg  -21} & {\bg  -19} &   -12 \\
\ru                      &                    & {\bg   -1} & {\bg  -13} &     0 \\ \cline{2-5}
\ru                      & \multirow{1}{*}{1} & {\bg  -22} & {\bg   12} &     5 \\ \cline{2-5}
\ru                      & \multirow{1}{*}{2} & {\bg   49} & {\bg   -6} &     5 \\ \cline{2-5}
\ru                      & \multirow{5}{*}{3} & {\bg    5} & {\bg   11} &     4 \\
\ru                      &                    & {\bg   11} & {\bg   14} &    -3 \\
\ru                      &                    & {\bg   29} & {\bg   24} &   -13 \\
\ru                      &                    & {\bg   44} & {\bg   17} &    -4 \\
\ru                      &                    & {\bg   57} & {\bg    8} &    -2 \\ \hline\hline
\ru \multirow{4}{*}{1-3} & \multirow{1}{*}{0} & {\bg  -36} &    23 & {\bg  -47} \\ \cline{2-5}
\ru                      & \multirow{2}{*}{2} & {\bg    9} &    -2 & {\bg  -17} \\ 
\ru                      &                    & {\bg   12} &     5 & {\bg  -14} \\ \hline\hline
\ru \multirow{5}{*}{2-3} & \multirow{1}{*}{1} &    -3 & {\bg  -18} & {\bg   10} \\ \cline{2-5}
\ru                      & \multirow{1}{*}{2} &     5 & {\bg   26} & {\bg  -27} \\ \cline{2-5}
\ru                      & \multirow{2}{*}{3} &    14 & {\bg   25} & {\bg   23} \\
\ru                      &                    &    -7 & {\bg   11} & {\bg   22} \\ \hline
\end{tabular}
}
\vspace{2mm}
\caption{Two different organization of the same dataset.
Left: only the largest component is used for classification, $G$=1.
Profile $p$=$\{1\}$ includes all vectors where the first component, $x_1$, is largest.
The profile is divided in two cones, $c$=0, including vectors with $x_1<0$, and $c$=1, including the others.
Right: the two largest components are used for classification, $G$=2.
Profile $p$=$\{$1-2$\}$ includes all vectors where $x_1$ and $x_2$ are the largest components.
The profile is divided in four cones, according to the sign of these components.
The largest $G$ components are shown in bold and color.}
\label{fig:dataset}
\end{figure}

\begin{figure}[t]
\centerline{
\includegraphics[width=2.8cm]{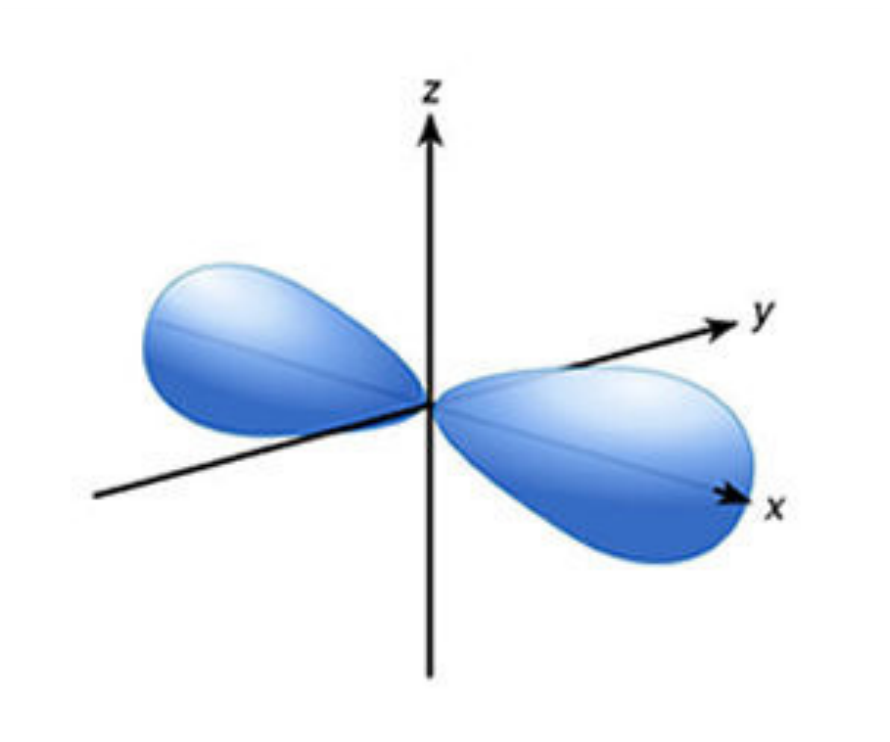}
\includegraphics[width=2.8cm]{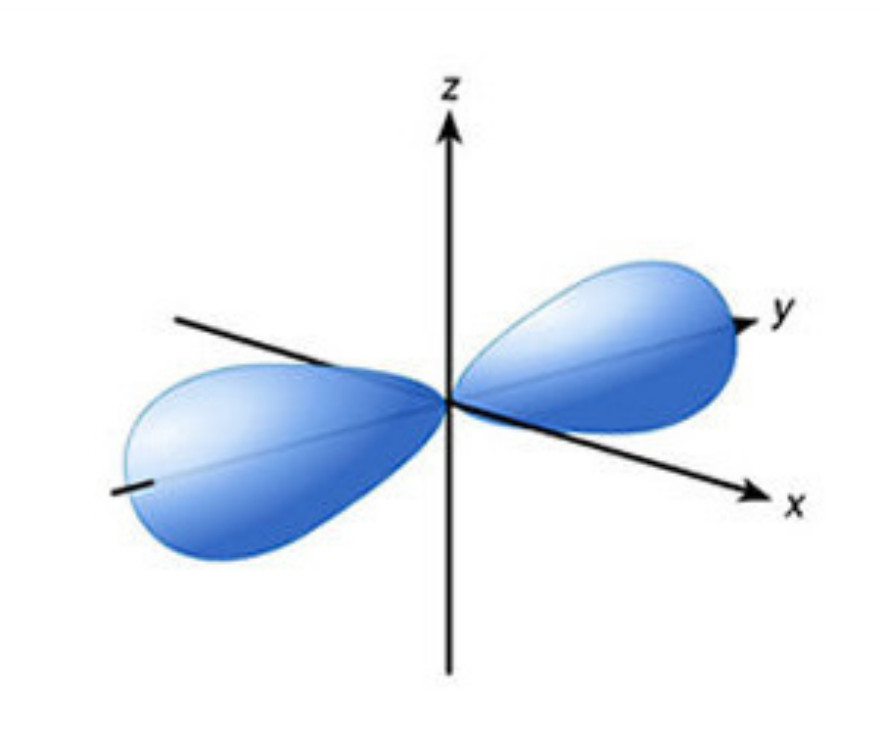}
\includegraphics[width=2.8cm]{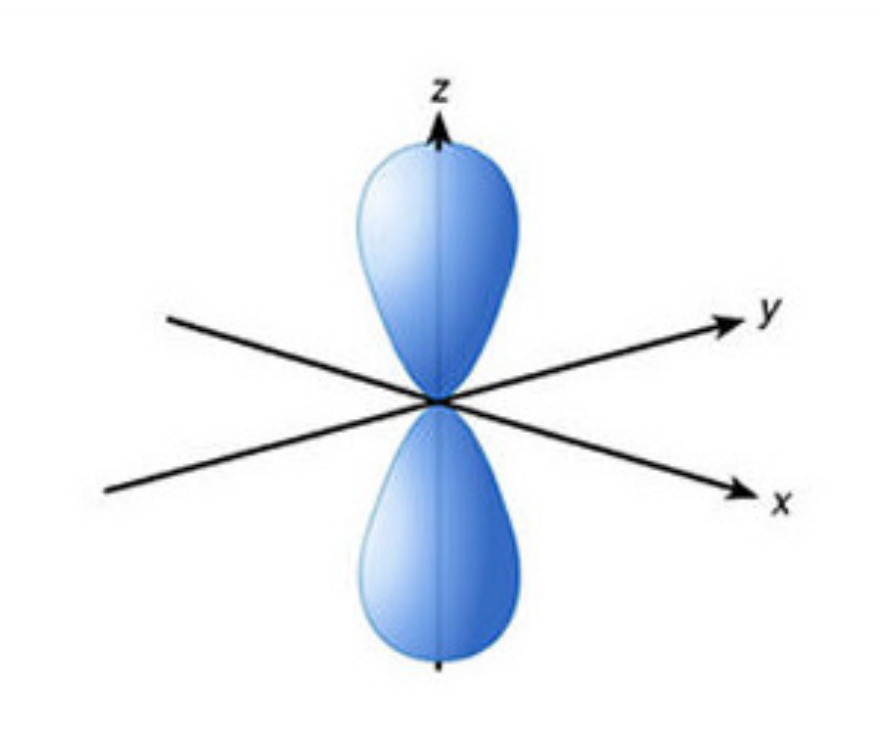}
}

\vspace{2mm}
\centerline{
\includegraphics[width=2.8cm]{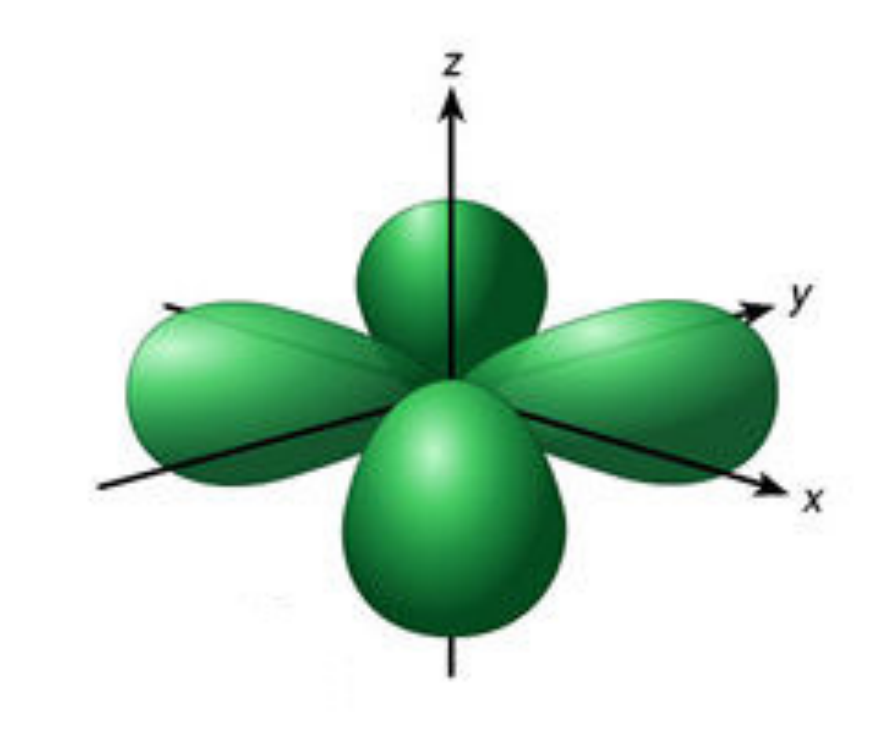}
\includegraphics[width=2.8cm]{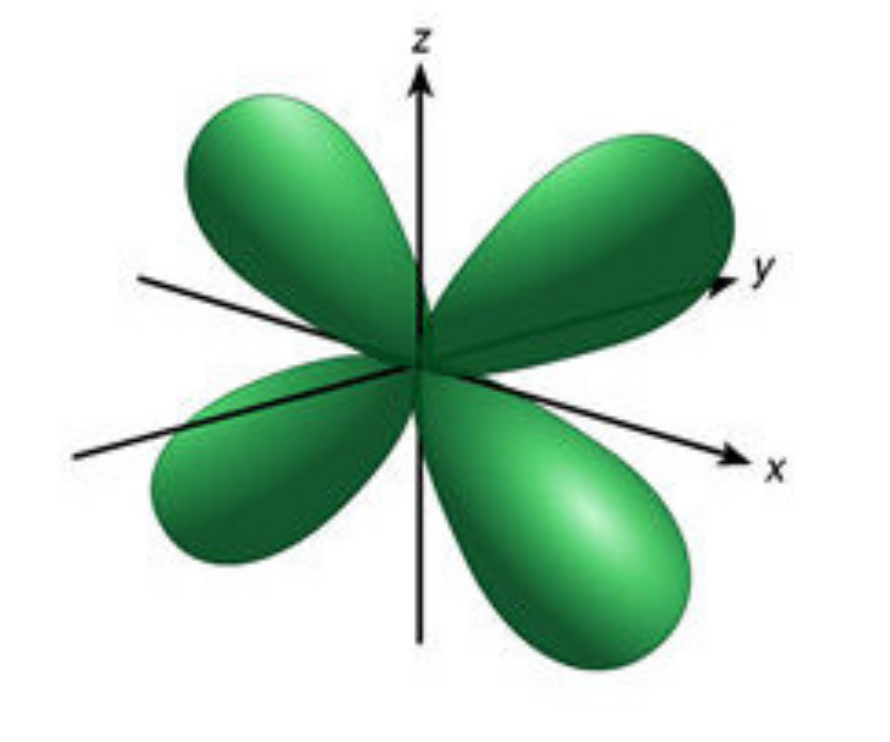}
\includegraphics[width=2.8cm]{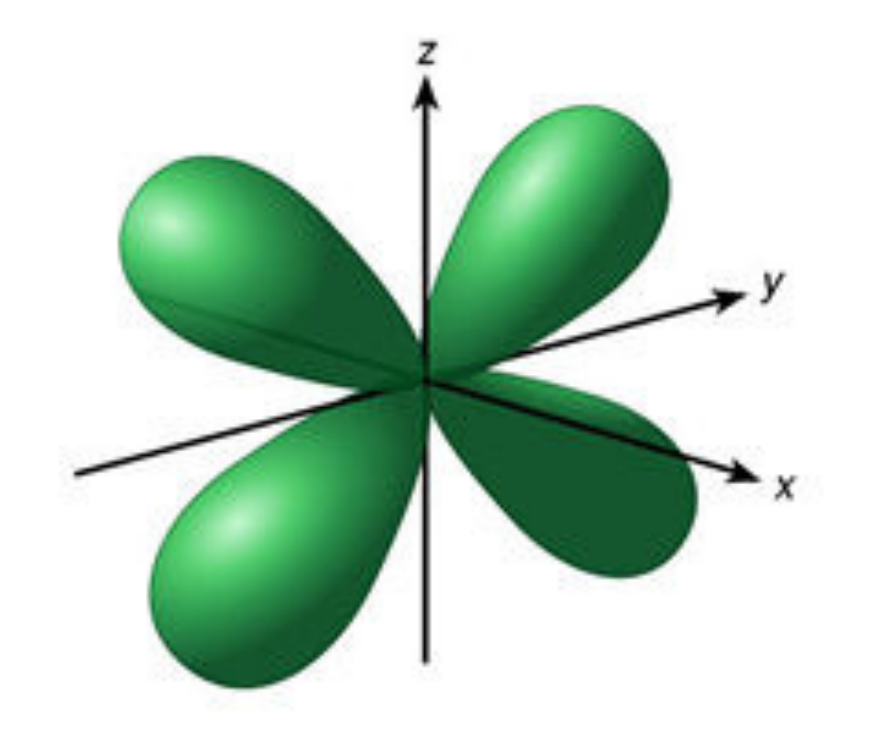}
}
\caption{Atomic orbitals $p_x, p_y$ and $p_z$ (top row) resemble (loosely) the three profiles arising in a 3d space with $G$=1, including two cones each.
Likewise orbitals $d_{xy}, d_{xz}$ and $d_{xz}$ (bottom row) resemble the three four-cone profiles of the case $G$=2.
The same color coding as figure 2 is used.
}
\label{fig:orbitals}
\end{figure}

Given a set of $N$ vectors $\x_n \in \RR^K, n=1$...$N$, drawn from a common source,
we look for the nearest neighbor to query $\y$ drawn from the same source, according to the Euclidean\footnote{The angular distance is also appropriate for ROSANNA,
but we focus on the Euclidean distance for its higher relevance in real-world problems.} distance,
\begin{IEEEeqnarray}{c}
    \| \x_{\rm NN}-\y\|^2 \leq \|\x_n-\y\|^2, \; n=1,\ldots,N
\end{IEEEeqnarray}

We organize in advance the dataset in disjoint sets, called profiles,
based on the {\em index} of the $G$ vector components that are largest in absolute value\footnote{In the following,
we omit ``in absolute value'' when obvious.}.
For example,
taking $G$=1, we define $K$ disjoint profiles, with profile $j$ including only the vectors for which the $j$-th component is the largest
\begin{equation}
    p_j = \{\x_n: |x_{nj}| \geq |x_{ni}|, \; i=1,\ldots,K\}
\end{equation}
Within each profile,
we further divide the vectors in subsets, called cones, according to the {\em sign} of the largest components, for example only two cones if $G$=1.
Fig.\ref{fig:dataset} shows two alternative organizations of a toy dataset, composed by 16 3d vectors, in the cases $G$=1 and $G$=2.
Taking advantage of some well-known pictures of atomic orbitals,
Fig.\ref{fig:orbitals} provides an approximate representation of profiles and cones in the 3d space, again for $G$=1 and $G$=2.

At run-time,
the query $\y$ is itself classified, based on index and sign of its $G$ largest components,
searching for the nearest neighbor only in the corresponding profile and cone.
In our classification, we tell apart the largest $G$ components from the $K$-$G$ smallest ones, but do not sort components within these groups.
Therefore, there are $\binom{K}{G}$ distinct profiles and, $N_C$ = $\binom{K}{G}2^G$ cones, counting $2^G$ cones for each profile.
In conditions of perfect symmetry for the source,
$N_C$ represents also the average speed-up, measured as the ratio between dataset size $N$ and number of vectors searched.
For the example dataset of Fig.\ref{fig:dataset}, $N_C$ equals 6 when $G$=1, and 12 when $G$=2.

This brief description elicits some natural questions:
is there potential for significant speed-up with this approach? Is the NN really likely to belong to the same cone as the query?
Both questions may have a positive answer when we move to high-dimensional spaces, $K \gg 1$,
thanks to the properties of order statistics.

\begin{figure}[t]
\centerline{
\includegraphics[width=4.2cm]{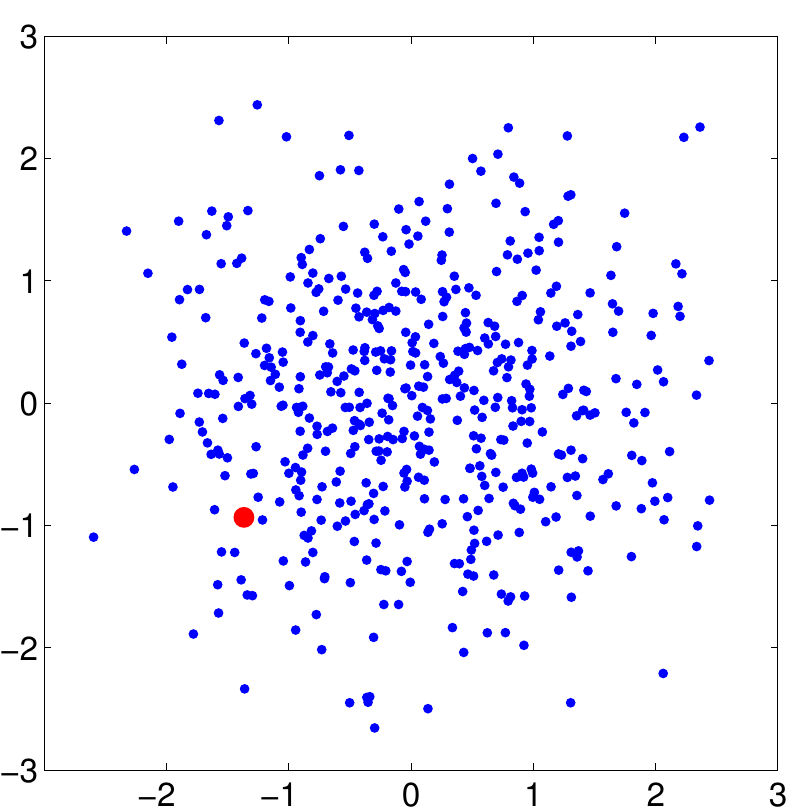}
\includegraphics[width=4.2cm]{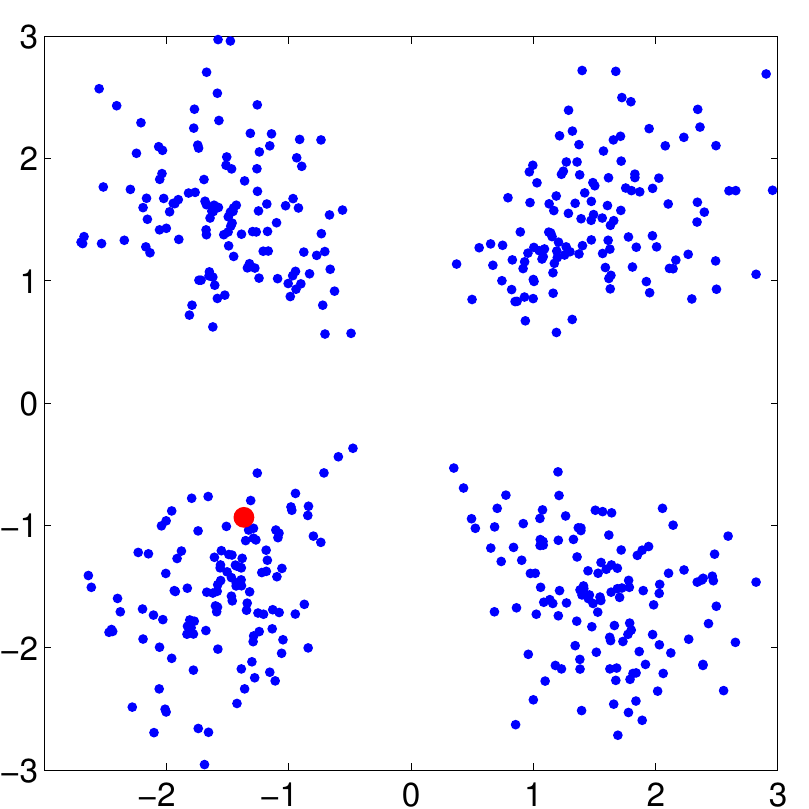}
}
\caption{Scatter plots of the first two components of 512 8d Gaussian vectors.
Left: randomly selected vectors. Right: vectors where the first two components are the largest ones.}
\label{fig:scatter_sort_nosort}
\end{figure}

Concerning speed-up,
it is important to note that the number of cones grows very quickly with $K$ and $G$.
For example, it is almost 30000 for $K$=16 and $G$=4.
Note that $K$=16 is considered a relatively small dimensionality for NN search problems.
By increasing the number of cones, one can reduce at will the average number of vectors per cone, hence the search time.
Needless to say,
in doing so, one should always guarantee a high probability that the NN is actually found in the searched cone (or cones).

To analyze this point,
let us focus on the specific case of vector components modeled as
independent identically distributed (i.i.d.) random variables with standard Gaussian
distribution, $X_i \sim {\cal N}(0,1)$.
This is arguably a worst case for the NN search problem, as there is no structure in the data to take advantage of.
It can be appreciated, for example, in the left scatter plot of Fig.\ref{fig:scatter_sort_nosort},
showing the first two components of 512 8-dimensional vectors with i.i.d. standard Gaussian components.
In the same plot we also show, in red, the first two components of a query drawn from the same distribution.
Clearly, there is not much structure in the data to help speeding up the NN search.
The right scatter plot of Fig.\ref{fig:scatter_sort_nosort} is obtained as the left one,
except that we now include only vectors such that the first two components are also the largest ones.
The difference between the two scatter plots is striking:
sorting the components has created a structure in the data, which can be exploited to speed-up the search.
In particular, we can safely restrict attention to only one of the four emerging clusters (our cones),
comprising vectors where the sign of the two largest components is the same as in the query, gaining therefore a factor four in speed.
Still, it is not impossible that the true NN belongs to a different cluster (it depends on the remaining components)
but there is an {\em energy gap} to overcome, due to the query-cluster distance in the first two components.

\begin{figure}[t]
\centerline{\includegraphics[width=8cm]{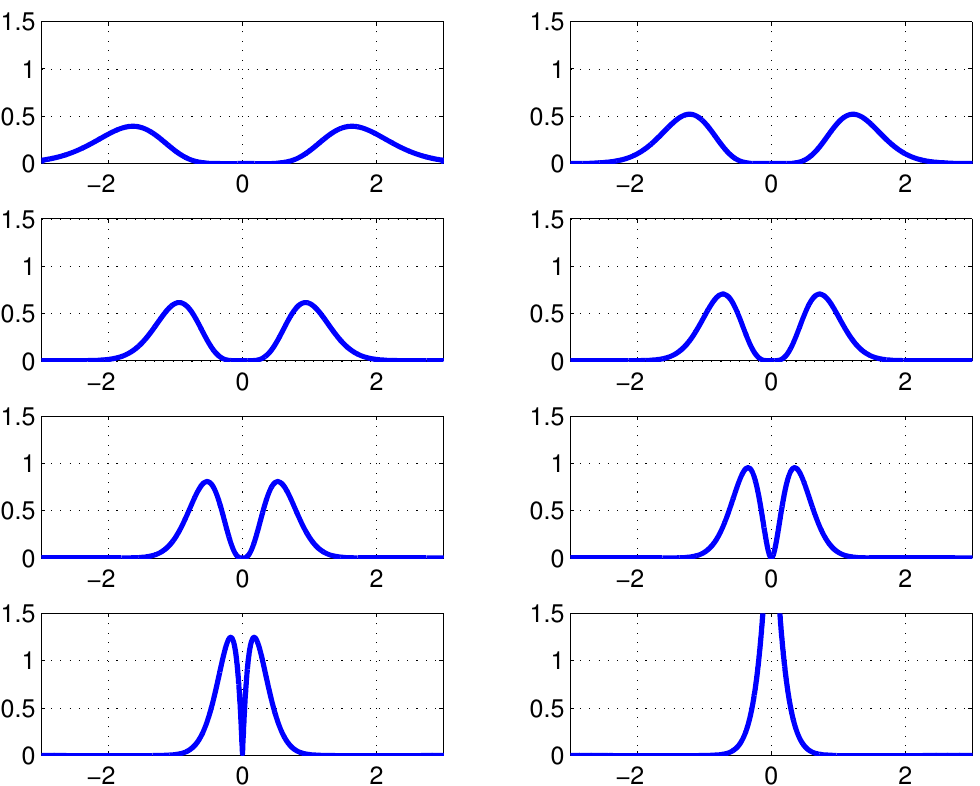} }
\caption{pfd of the components of an 8-dimensional vector of i.i.d. standard Gaussians after sorting for decreasing magnitude.
The largest components (top) have a high variance and are clearly bi-modal.
The first component alone holds 44\% of the vector energy, the first four almost 90\% of it.}
\label{fig:Gaussian_OS_pdf}
\end{figure}

Of course, after sorting,
the components are not identically distributed anymore, and certainly not independent on one another.
Given the probability density function (pdf) of the original components, $f_X(x)$, we can easily compute the pdf of the sorted components.
Let $A_i=|X_i|$ be the absolute value of the $i$-th component,
and $A_{(i)}$ the $i$-th component of the sorted vector of absolute values, such that $A_{(i)} \geq A_{(i+1)}$.
Note that $A_{(1)}$ is smaller than a given value $x$ only if all the $A_i's$ are.
Likewise, $A_{(i)}$ is smaller than $x$ only if {\em at least} $K \minus i \plus 1$ of the $A_i's$ are.
Based on such observations, and due to the independence of the $A_i's$,
we can compute the marginal pdf of the sorted absolute values (see (\ref{eq:OS_pdf_1}), Appendix A)
\begin{IEEEeqnarray}{rCl}
    f_{A_{(i)}}(x) & =      & \frac{K!}{(K-i)!(i-1)!} \nonumber \\[1mm]
                   & \times & [F_A(x)]^{K-i} [1-F_A(x)]^{i-1} f_A(x)
    \label{eq:OS_modulus_pdf}
\end{IEEEeqnarray}
where $F(\cdot)$ denotes cumulative distribution function (CDF).
Then, given (\ref{eq:OS_modulus_pdf}), we readily obtain the pdf of the original components after sorting them by decreasing magnitude.

In our example,
the components are standard Gaussian, with CDF expressed in terms of the $Q$-function \cite{Papoulis2002} as $F_{X_i}(x)=1-Q(x)$.
For the case $K$=8,
Fig.\ref{fig:Gaussian_OS_pdf} shows the pdf of all components, which are very different from one another.
The first components (top) have a much larger variance than the last ones, holding most of the vector energy.
Therefore,
they impact heavily in the computation of the Euclidean distance w.r.t. a given query,
while the last ones are almost negligible.
Moreover, the largest components are markedly bimodal, with modes growing farther apart as $K$ grows.

This figure provides, therefore, some more insight into the rationale of our approach.
We are trying to classify vectors beforehand, in a sensible way, to reduce the search space.
Doing this by taking into account {\em all} components with equal importance would be impractical (or infeasible) as $K$ grows large,
and not much reliable, because most components are scarcely informative.
Therefore, we focus  only on the largest components, those holding most of the energy and of the information content,
obtaining a much smaller (and tunable) number of classes and, eventually,
some stronger guarantee that the NN will indeed belong to the same profile as the query.
As a matter of fact, we chose the name ``profile'' for analogy with the actions naturally taken to identify a person based on a summary description,
focusing on the most prominent features,
{\it ``...he had the most unusual nose...'', ``...she had a curious accent...''},
to reduce the search space while preserving accuracy.
Given the profile, and assuming the NN is actually found in that class,
the analysis of signs restricts the search very reliably on the cone of interest.
In fact, since the largest components have such a strongly bimodal distribution,
it is very unlikely that the smallest components cause a cone switch.

\begin{figure}[t]
\centerline{
\includegraphics[width=4.2cm]{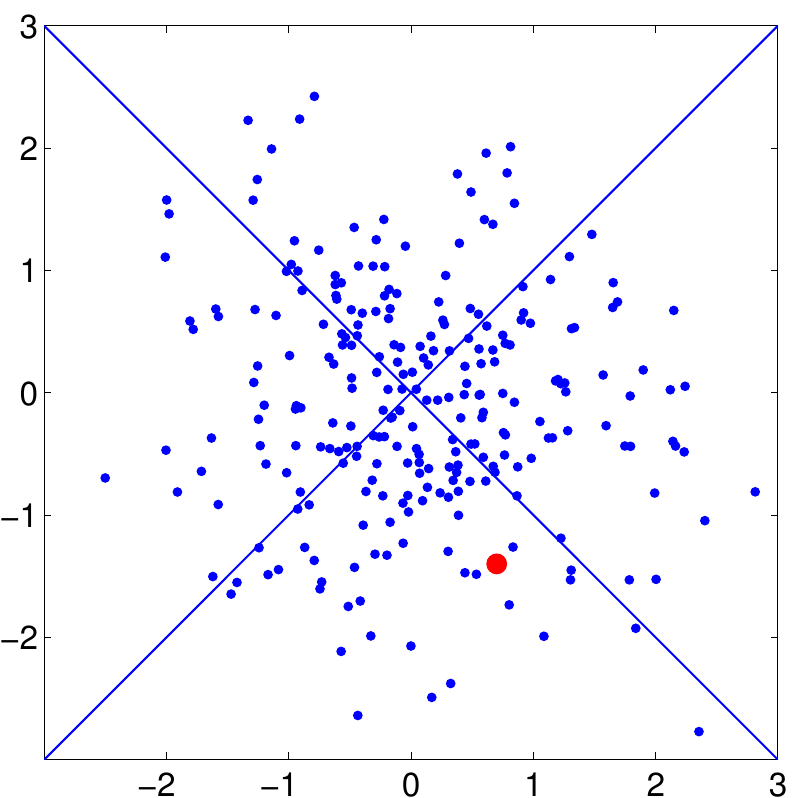}
\includegraphics[width=4.2cm]{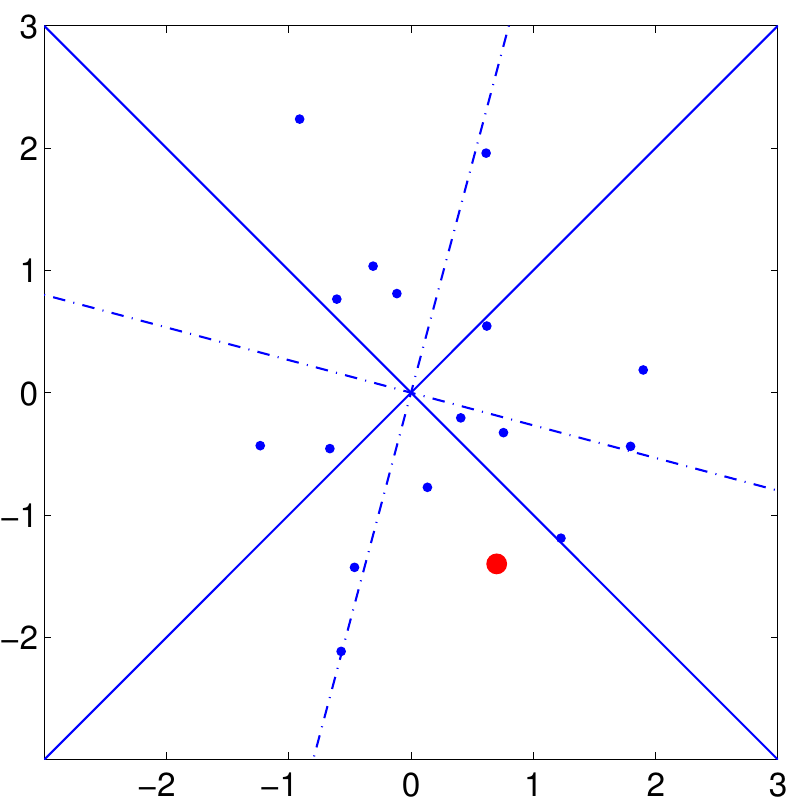}
}
\caption{Scatter plots of 2-dimensional i.i.d. Gaussian vectors.
At very high density (left, $\rho$=4) the NN belong almost always to the query's cone.
This may not happen at lower density (right, $\rho$=2), especially if the query is near to the cone boundaries.
A rotation of coordinates brings the query near the center of the new cone (dash-dot lines),
with the NN in the same cone.
}
\label{fig:lowhigh density}
\end{figure}

Taking a different point of view, ROSANNA can be seen as a form of locality sensitive hashing.
Component sorting becomes just a means to determine algorithmically a partition of the space based on vector direction.
Given the identity of the $G$ out of $K$ largest components,
a data vector is automatically associated with one of the cells of the partition, and the same happens with the query.
With $G$=1, the space in divided in 2$K$ cells, our cones,
which become 2$K(K-1)$ with $G$=2, and so on, up to $2^K$ cells for $G$=$K$.
It is worth underlining that the space partition is, by definition, completely symmetric,
and induces a partition of the unit hyper-sphere with the same property.
In Appendix B we characterize the proposed approach in terms of collision probability.

\section{Implementation}

\begin{algorithm}
\footnotesize
\begin{spacing}{1.1}
\begin{algorithmic}
   \Require $\y$	                                       \Comment query
   \Ensure NN	                                           \Comment index of approximate nearest neighbor
      \For{$r = 1:R$}                                      \Comment for each rotation
      \State compute $\y^{(r)}$                            \Comment projection of $\y$ on $r$-$th$ basis
      \State find $\{c_1^{(r)},\ldots,c_C^{(r)}\}$         \Comment ordered list of cones to be visited
   \EndFor
   \For{$l = 1:C$}                                         \Comment $C$ cones visited for each basis
      \For{$r = 1:R$}
	     \For{each $x \in c_l^{(r)}$}
		    \If {$x$ not analyzed}                         \Comment boolean side information
			   \State compute $\parallel \x-\y\parallel^2$ \Comment with partial distance elimination
			   \State update NN
			   \State mark $x$ as analyzed
            \EndIf
         \EndFor
      \EndFor
   \EndFor
\end{algorithmic}
\end{spacing}
\caption{ROSANNA (NN search)}
\label{algo:ROSANNA}
\end{algorithm}

We now turn the naive basic algorithm into a {\em reliable} ANN search tool.
The weak point in the basic version
is the assumption that the NN is found {\em exactly} in the cone singled out by the query.
This is quite likely if the dataset has a high density of points,
\begin{equation}
    \rho = \log_2N/K
\end{equation}
as in the 2d example on the left of Fig.\ref{fig:lowhigh density}, much less so in the case of lower density, shown on the right.
This latter plot shows how the NN may happen not to be in the query's cone,
especially if the query lies near the boundary of the cone and not in the very center.
This might look as a rare unfortunate case.
However, in high-dimensional spaces, this is actually quite likely, especially at low density.
As a matter of facts, even in the right plot the point density is actually quite high,
while in most real-world applications densities in the order of $\rho$=1 or even lower are to be expected,
in which case the NN may easily happen to be far from the query.
Fig.\ref{fig:lowhigh density}, however, suggests also possible countermeasures,
amounting basically in considering alternative bases (see the dash-dot lines on the right), obtained through rotation,
or including also neighboring cones in the search.

To make our OS-based search reliable we resort therefore to some typical expedients of LSH methods,
enlarging the set of candidate points and exploring them with suitable priority.

\subsection{Using multiple bases}
To increase the reliability of our search algorithm we deal first with the boundary problem.
Although this is not obvious in the 2d case,
in higher-dimensional spaces it is quite likely that the query lies far from the center of its cone.
When this happens, the probability that the NN belongs to a different cone is quite large, exceeding 1/2 when the query lies exactly on a boundary.
To address this problem, we consider multiple reference systems, obtained from one another through random rotations, like in \cite{Anan2008},
and look for the NN in the union of all the cones where the query belongs.
This solution corresponds to the use of multiple hash tables in LSH algorithms, and presents the same pros and cons.
The probability of finding the NN in the enlarged cone is much higher than before,
but there is a processing cost, since the query is projected on multiple bases and more points are checked,
and a memory overhead, due to the need to store multiple classifications.

\subsection{Checking neighboring cones}
Using multiple bases increases the probability of finding the NN in the query's enlarged cone,
but there is still a non-negligible probability of missing it, especially in the low-density case.
Therefore, it can make sense to extend the search to some close cones, as far as a positive time-accuracy trade-off is kept,
which is the multiprobe search used in LSH methods \cite{Lv2007}.
Rather than computing the actual Euclidean distances between the query and candidate cones,
we exploit the intrinsic structure of profiles defined for various values of $G$.
Let $i_1, i_2, \ldots, i_K$ be the indexes of the query coordinates sorted by decreasing magnitude.
Therefore, for a given value $G$, the query belongs to the profile identified by $\{i_1,i_2,\ldots,i_G\}$.
The most likely reason why the NN may not belong to the same profile is that its $G$-th coordinate differs from the query's,
so we should begin by looking in all profiles sharing the first $G \minus 1$ indexes with the query's profile but differing in the last one.
Note that all such level-$G$ profiles are ``children'' of the same level-$(G \minus 1)$ profile,
identified by $\{i_1,i_2,\ldots,i_{G \minus 1}\}$.
Therefore, we define a distance between two level-$G$ profiles, $p_1$ and $p_2$, as $G$ minus the level of their closest common ancestor.
Based on such a distance, for each rotation an ordered list of cones to be visited is established.

A compact simplified pseudo-code of ROSANNA (only the search phase) is shown in Algorithm \ref{algo:ROSANNA}.

\subsection{Preliminary assessment of reliability}

We ran a few experiments to gain insight into the importance of using multiple bases and searching neighboring cones,
and to assess the reliability of the proposed search algorithm.

\begin{figure}[t]
\centerline{
\includegraphics[width=4.2cm]{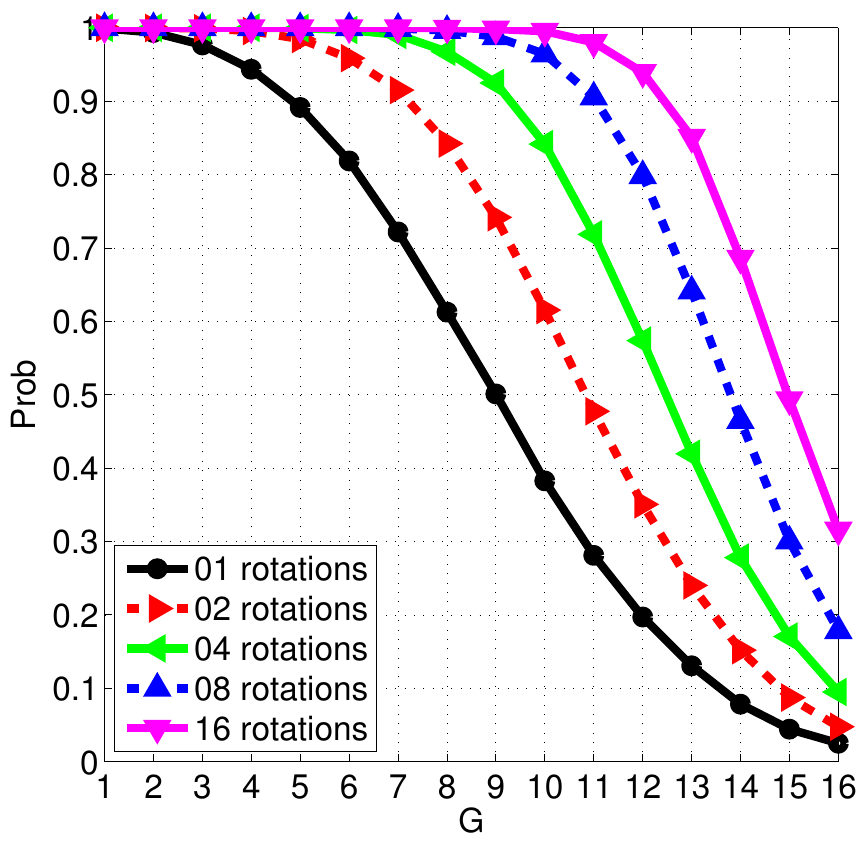}
\includegraphics[width=4.2cm]{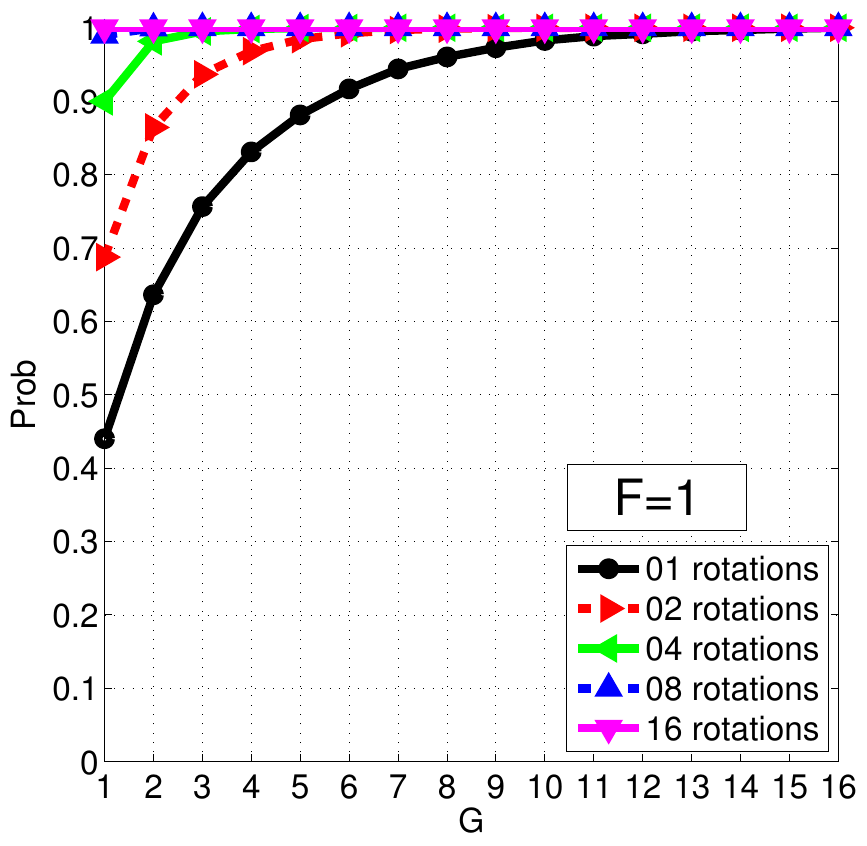}  }

\centerline{
\includegraphics[width=4.2cm]{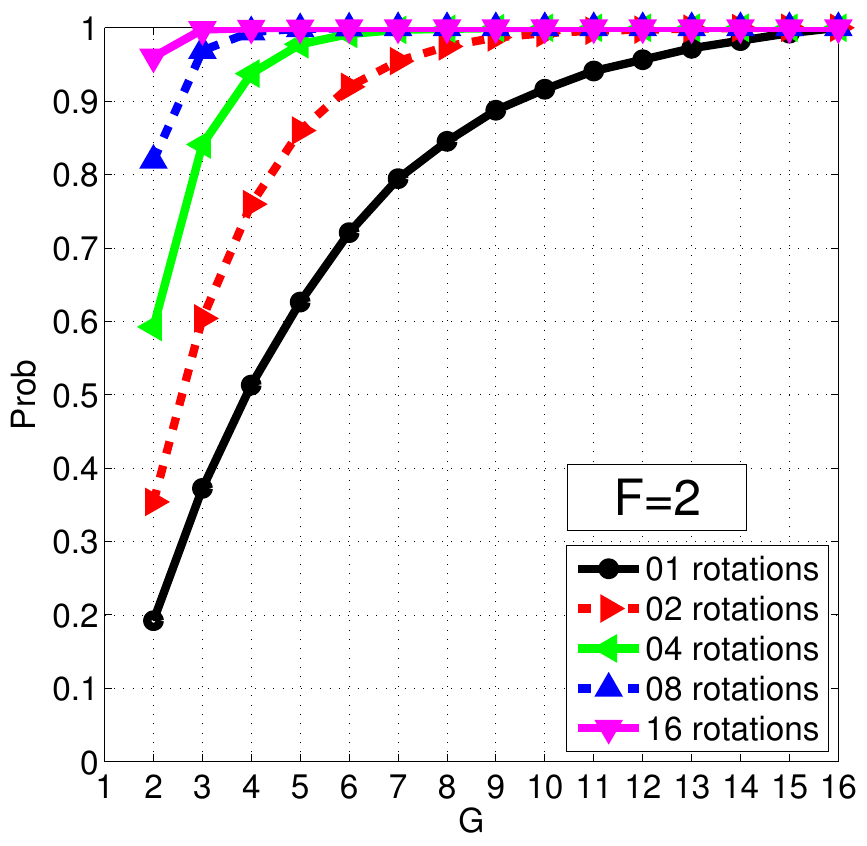}
\includegraphics[width=4.2cm]{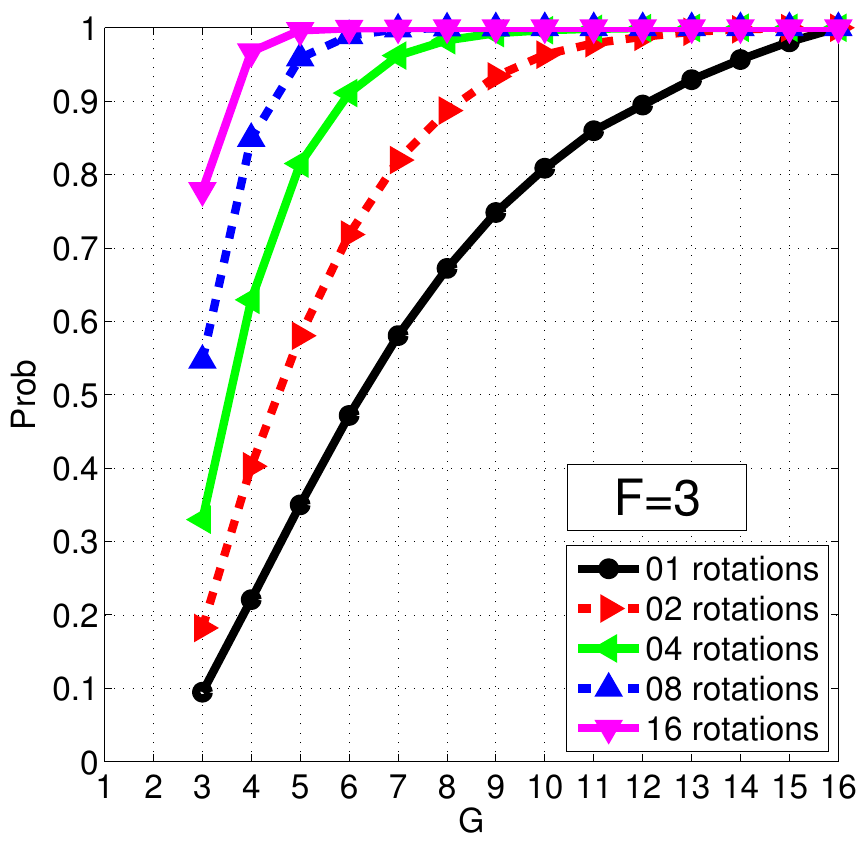}  }
\caption{i.i.d. Gaussian data, $\rho$=1.
Sign agreement between query and NN (top-left) is quite high, especially with multiple rotations.
Classification accuracy (other graphs), instead, is guaranteed only with a small number of query components, $F$, and many rotations. }
\label{fig:preliminary1}
\end{figure}

We consider a running example with i.i.d. Gaussian components, with $K$=16 and $\log_2N$=16, hence density $\rho$=1,
and report in Fig.\ref{fig:preliminary1} (top-left) the probability (MonteCarlo estimates)
that the first $G$ components of query and NN have the same sign.
Using a single basis, this probability is very large only for small values of $G$:
for example,
while it is almost certain that the first 3 components have the same sign, the probability drops to about 60\% for $G$=8.
Note that for $G$=16, the probability is almost 0,
showing that using signs for classification {\em without} prior sorting, like in ITQ, is prone to errors.
Using multiple bases guarantees a significant improvement,
and already with 8 rotations the first 8 components of query and NN have almost certainly the same sign.

The top-right plot of Fig.\ref{fig:preliminary1}
reports the probability that the largest component of the query ($F$=1) is among the $G$ largest components of the NN.
The probability is just above 40\% for $G$=1,
that is, even in this relatively high-density case, we cannot fully trust the largest component for reliable classification.
If we look also in profiles where the largest component of the query is only the second, third, ..., $G$-th largest,
the probability of finding the NN grows, but not as quickly as we might hope.
However, by using multiple bases, the whole curve drifts rapidly towards 1, becoming almost flat for 8 rotations.
To increase search efficiency, however,
classification should be carried out based on more components, $F>1$.
The bottom plots report, therefore, the same data with reference to the two and three largest components of the query, respectively.
In both cases, the classification is very reliable for small values of $G$ if 8-16 rotations are used.

\begin{figure}[t]
\centerline{
\includegraphics[width=4.2cm]{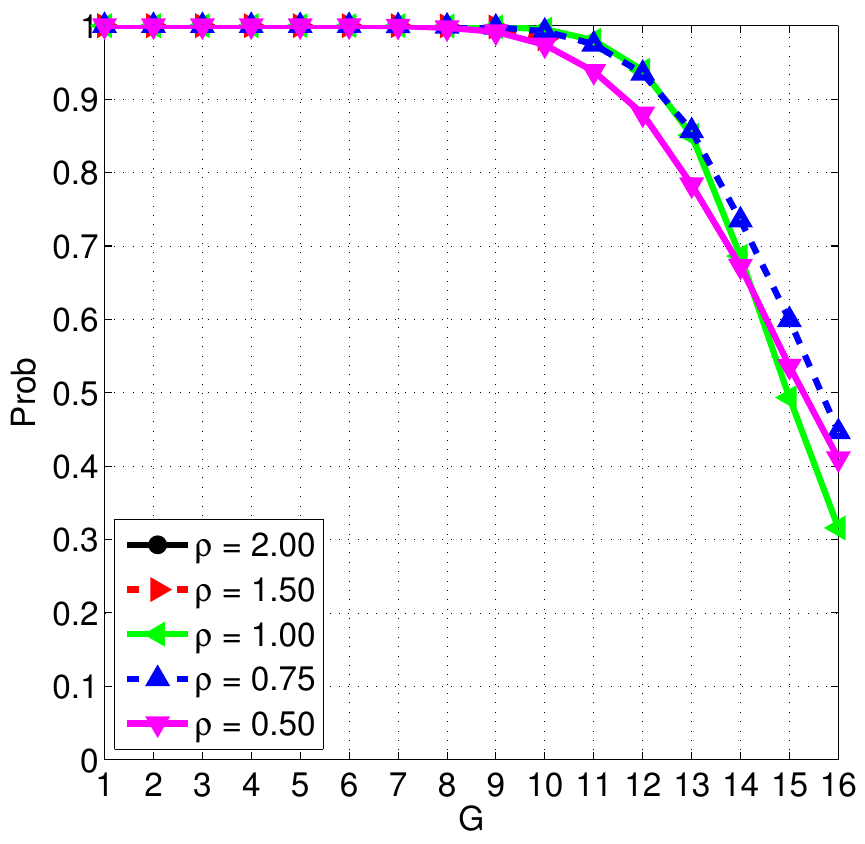}
\includegraphics[width=4.2cm]{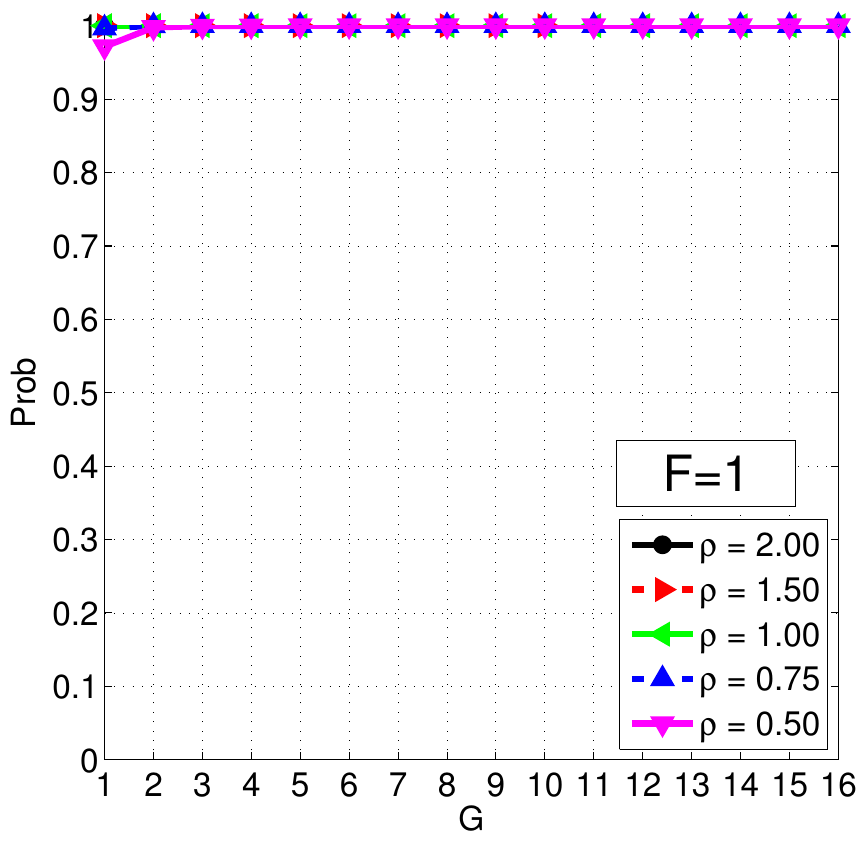}   }

\centerline{
\includegraphics[width=4.2cm]{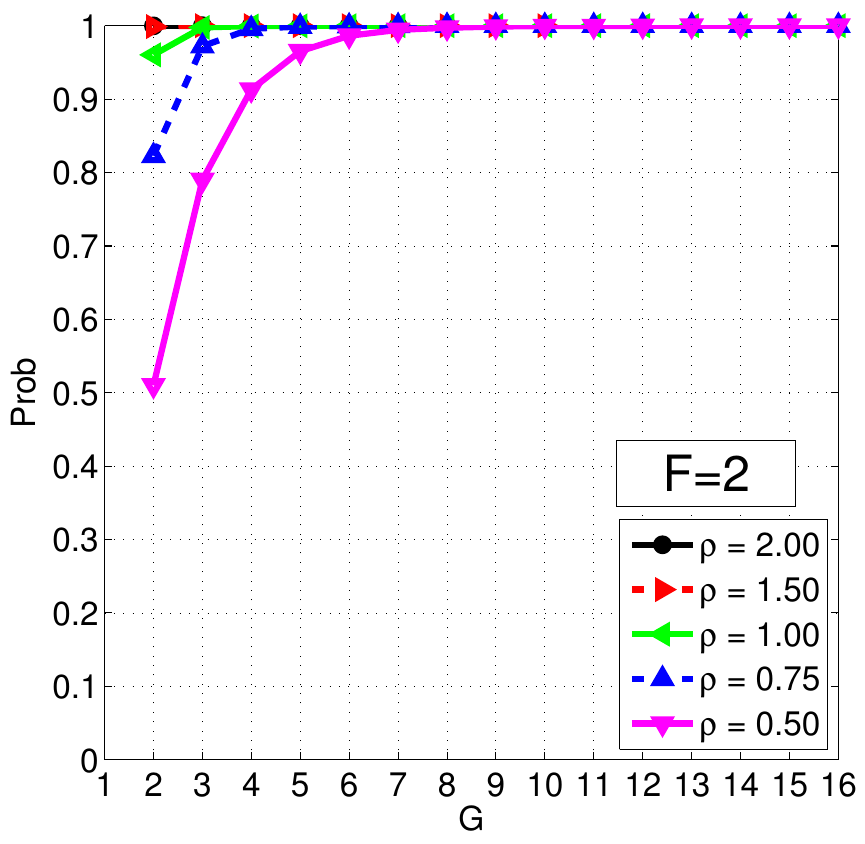}
\includegraphics[width=4.2cm]{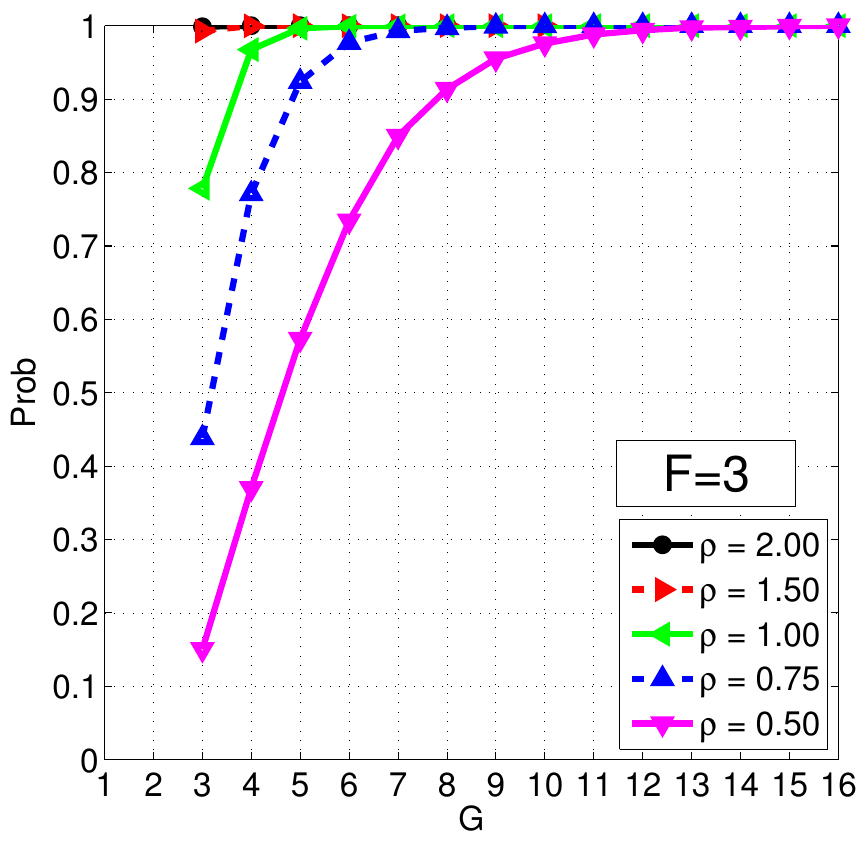}   }
\caption{i.i.d. Gaussian data, 16 rotations.
Sign agreement between query and NN (top-left) is high at all tested densities.
Classification accuracy (other graphs) is limited at low density if more than two query components are used. }
\label{fig:preliminary2}
\end{figure}

\begin{figure*}[t]
\centerline{
\includegraphics[width=5.6cm]{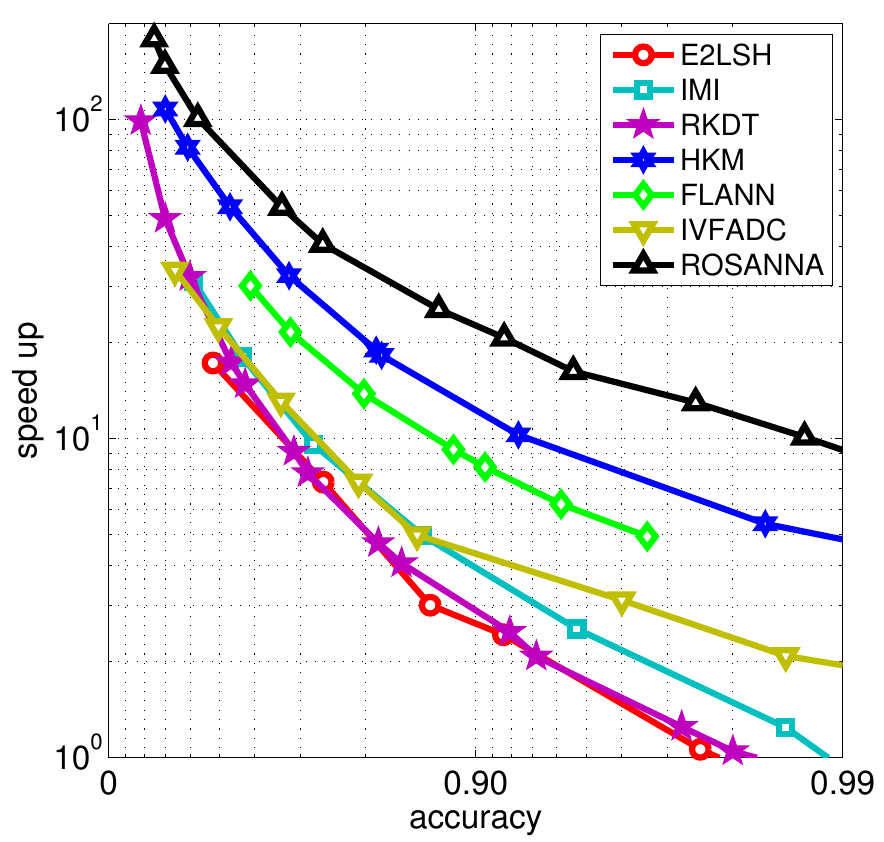}
\includegraphics[width=5.6cm]{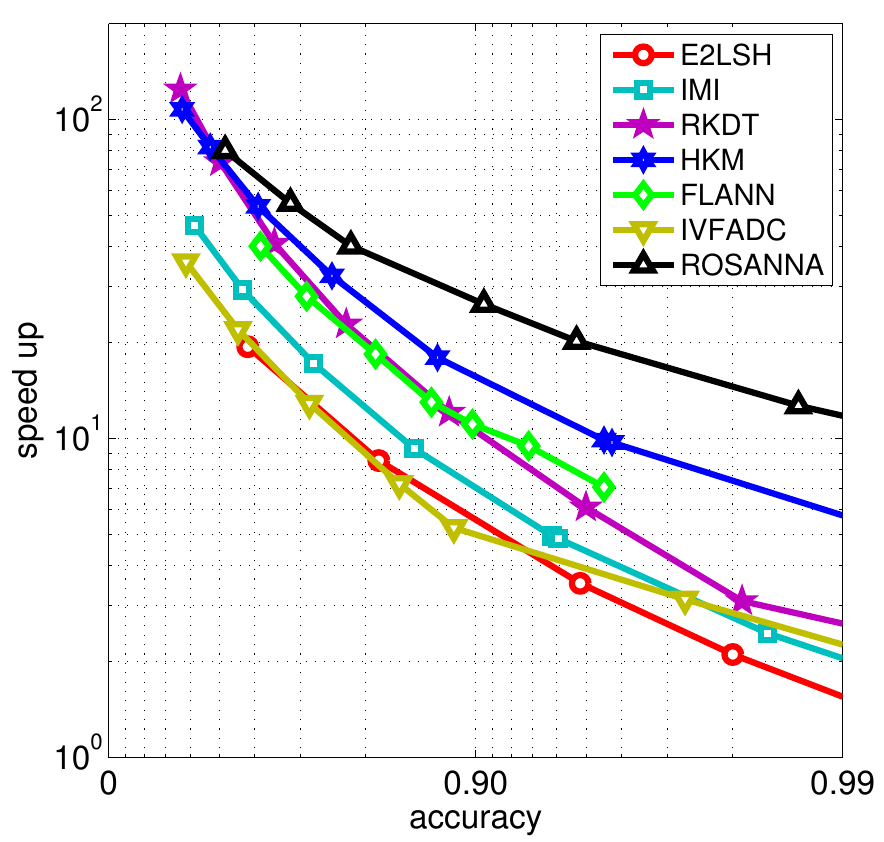}
\includegraphics[width=5.6cm]{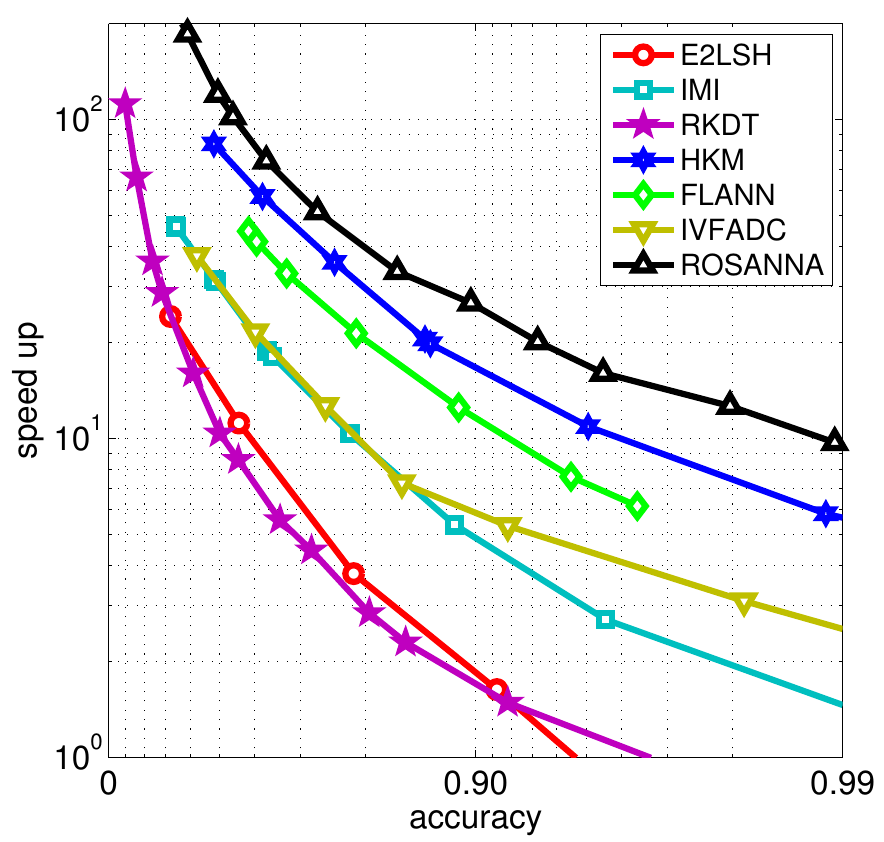}
}
\caption{Experimental results at $\rho$=1 with Gaussian (left), Uniform (center), and Laplace (right) i.i.d. data.
In all cases ROSANNA, outperforms all references, sometimes by an order of magnitude, and results are only weakly affected by the data pdf.}
\label{fig:rho100}
\end{figure*}

In Fig.\ref{fig:preliminary2} we explore the dependence on dataset density, going from 2 (high) to 1/2 (quite low), with 16 rotations.
The density is modified by keeping $\log_2N \simeq$ 16 and varying the vector length $K$.
In any case, results are shown for $G \leq 16$, for clarity.
The top-left plot shows that sign agreement between query and NN is always guaranteed, at least for the first 8 components.
The other plots report the probability that the largest 1, 2, 3 components, respectively, of the query are among the $G$ largest of the NN.
This probability is very high for $F$=1 at any density, and for any $F$ at high density.
At low density, instead, classification becomes unreliable for $F>2$ unless a large value of $G$ is used,
which is impractical, however, as it would require visiting a huge number of profiles.
Therefore, we cannot expect the search to be much reliable in this case.
On the other hand, the same holds (to the best of our knowledge) for all other ANN search methods at such low densities.
Indeed,
most large-dimensionality sources considered in the literature
present strong dependencies which reduce significantly the intrinsic data dimensionality \cite{Jegou2012, Gong2013, Ge2014a},
allowing for an effective ANN search.

\subsection{Assessment of complexity}
We can now provide a theoretical assessment of computational complexity,
keeping in mind, however, that some processing steps include random components that may affect results significantly.
To this end, Tab.\ref{tab:algorithm_parameters} lists the main parameters of the algorithm and the associated symbols,
while Tab.\ref{tab:complexity} reports the complexity assessment as a function of these quantities.

\begin{table}[t]
\centering
{\footnotesize
\begin{tabular}{cl}\hline
\ru $N$   & number of dataset vectors                 \\
\ru $K$   & vector length                             \\
\ru $G$   & number of components used for query       \\
\ru $N_C$ & number of cones, $N_C=\binom{K}{G}2^G$    \\
\ru $R$   & number of rotations (hash tables)         \\
\ru $C$   & number of cones visited for each rotation \\ \hline
\end{tabular}
}
\vspace{2mm}
\caption{Main parameters of the algorithm.}
\label{tab:algorithm_parameters}
\end{table}

\renewcommand{\ru}{\rule{0mm}{3mm}}
\renewcommand{\O}[1]{{O}(#1)}
\begin{table}[t]
\centering
{\footnotesize
\begin{tabular}{|c|l|l|}\hline
\ru phase                        & action                     & complexity         \\ \hline
\ru                              & generate rotation matrices & $\O{RK^3}$         \\
\ru dataset preparation          & rotate dataset points      & $\O{NRK^2}$        \\
\ru                              & hash dataset points        & $\O{NRK\log(K)}$   \\ \hline
\ru                              & rotate query               & $\O{RK^2}$         \\
\ru NN search                    & hash query                 & $\O{RK\log(K)}$    \\
\ru                              & search short-list          & $\O{RCK(N/N_C)}$   \\ \hline
\end{tabular}
}
\vspace{2mm}
\caption{Complexity assessment.}
\label{tab:complexity}
\end{table}

The dataset preparation phase is normally of minor interest since it is carried out off-line once and for all.
Considering that $N$ is much larger than $K$, the dominant term of this phase is the rotation of the dataset points along the $R$ bases.
We are considering the use of structured orthonormal matrices, like DCT or Walsh-Hadamard, in place of pseudo-random matrices, which may reduce this cost.
Hashing, instead, has  more limited cost, related to vector sorting.

For on-line NN search the most critical item is typically the linear search of the candidate short-list where the distance from query to all candidates must be computed.
However, the corresponding entry in Tab.\ref{tab:complexity} is only an approximation,
based on the assumptions that dataset points are uniformly distributed among the cones, and that the visited cones include disjoint sets of points.
The first assumption is pretty reasonable, the second much less.
When multiple rotations are used, it is very likely that some points are visited more than once, in which case the distance is not computed anew.
Therefore, our estimate is a bit pessimistic, but how much so depends on many parameters.
We can however single out best and worst cases.
In the best case,  the shortlist includes only one candidate, and search complexity is dominated by the cost of query rotation $\O{RK^2}$.
In the worst case, the shortlist includes all dataset points, coming down to linear search, with cost $NK$.

In next Section,
ROSANNA will be applied also to long vectors (e.g., 128 components) reduced to shorter unstructured vectors through PCA and random rotation.
Taking into account also the estimation of covariance matrix and the PCA, and the use of partial distance elimination techniques,
the above analysis holds with minor adjustments also in such a case.

\section{Experimental analysis}

\begin{figure*}[t]
\centerline{
\includegraphics[width=5.6cm]{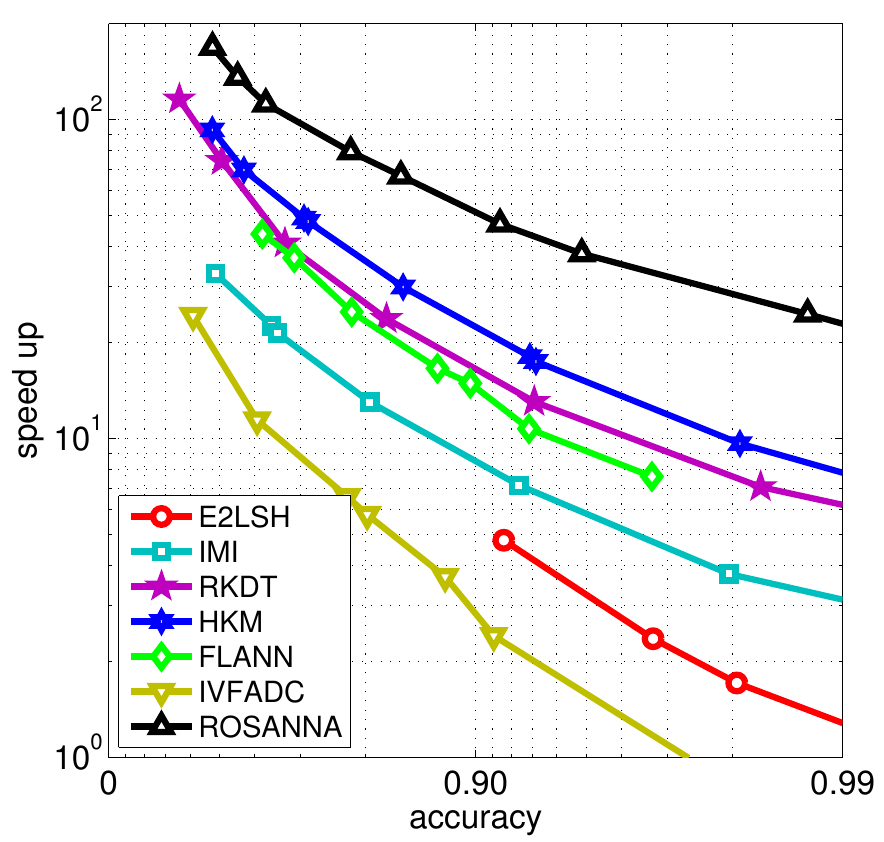}
\includegraphics[width=5.6cm]{figure/N_rho100_comparison}
\includegraphics[width=5.6cm]{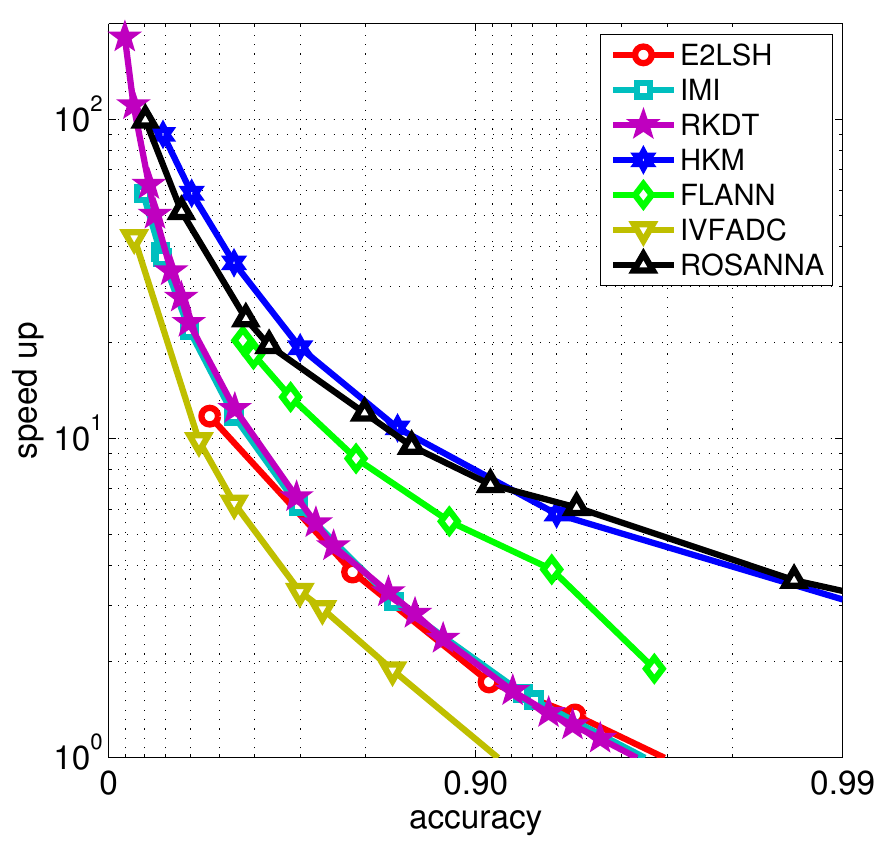}
}
\caption{Experimental results with Gaussian i.i.d. data at high ($\rho$=1.5, left), medium ($\rho$=1, center), and low ($\rho$=0.75, right) density.
Performance depends strongly on density.
ROSANNA works much better than the references at high density, and is on par with HKM at low density.
In the latter case, a very limited speed-up is obtained anyway.
}
\label{fig:Gaussian_rhovar}
\end{figure*}

We carried out a number of experiments to assess the performance of the proposed method.
Results are given in terms of accuracy-efficiency plots, as in \cite{Muja2014, Iwamura2013}
with accuracy meant as the probability that the selected point is the actual NN (also called precision and recall@1),
and efficiency measured in terms of speed-up w.r.t. linear search.
The software is written in C++ using open libraries and some routines of the FLANN package\footnote{\url{http://www.cs.ubc.ca/research/flann/}},
and is published online\footnote{\url{http://www.grip.unina.it}} to guarantee full reproducibility of results.
There are only a few parameters to set:
the number of components used for classification, $G$,
the number of rotations, $R$, used for multiple-basis search,
and the number of visited cones per basis, $C$.
In the preparation phase,
for each rotation, all dataset points are projected on the new basis and classified according to the index and sign of their $G$ largest components.
Therefore we need $R$ hash tables, with a relative memory overhead of $R/K$.
A second-level hash table is used to manage the tables when the number of cones grows very large.

We compare results with a number of relevant state-of-the-art references:
{\it   i)} plain Euclidean LSH (E2LSH) \cite{Andoni2006} with the implementation available online\footnote{\url{http://www.mit.edu/~andoni/LSH/}} including the automatic setting of most parameters;
{\it  ii)} randomized kd-trees (RKDT) and,
{\it iii)} hierarchical k-means (HKM), both implemented in the FLANN package \cite{Muja2014}, together with
{\it  iv)} FLANN itself, which is always inferior to the the best of RKDT and HKM but sets automatically all parameters;
{\it   v)} the IVFADC algorithm \cite{Jegou2011}, based on product quantization, and implemented by us starting from the Matlab code
published by the authors\footnote{\url{http://people.rennes.inria.fr/Herve.Jegou/projects/ann.html}}, and finally
{\it  vi)} Inverted Multi-Index (IMI) \cite{Babenko2012} developed by the authors\footnote{\url{http://arbabenko.github.io/MultiIndex/}}
    except for the final linear search among candidates, which we carried out as in ROSANNA.
All these techniques are implemented in C++ language.
We also implemented and run optimized PQ, both parametric and non-parametric \cite{Ge2014a},
but did not include results, generally worse than those of the PQ-based IMI and IVFADC, in order not to clutter further the figures.
Curves are obtained (except for FLANN) as the upper envelope, in the accuracy-time plane, of points corresponding to different parameter settings.
For ROSANNA, we consider $G \in \{1,2,\ldots,K/2\}, R \in \{1,2,4,8,16\}, C \in \{1,2,4,\ldots,128\}$,
and similar wide grids are explored for the main parameters of all other techniques.
To focus on the more interesting high-accuracy range, in all graphs we use a logarithmic scale for both accuracy and speed-up.

\subsection{Unstructured data}

Fig.\ref{fig:rho100} (left) shows results for our running example, Gaussian i.i.d. data, $k$=16, $\rho$=1.
ROSANNA guarantees uniformly the best performance,
being almost twice as fast than the second best, HKM, at all levels of accuracy,
and much faster than all the other references, gaining a full order of magnitude w.r.t. RKDT and E2LSH.
The FLANN curve lies somewhat below HKM, since the parameters are selected in advance and may turn out not to be the best possible.
However, we could not compute results for FLANN at high accuracy due to the large time needed to optimize the parameters.
In Fig.\ref{fig:rho100} we also shows results obtained in the same conditions as before
but using Uniform (left) and Laplace (right) random variables in place of Gaussian.
The general behavior is the same as before,
with slight improvements observed in the Uniform case, probably due to the smaller entropy.

With Fig.\ref{fig:Gaussian_rhovar} we go back to the Gaussian case, but change the dataset density
considering $\rho=$1.5 (left), $\rho=$1.0 (center) as before, for ease of comparison, and $\rho=$0.75 (right).
As expected, ROSANNA works especially well at high density,
while its performance becomes very close to the best reference, HKM, at lower density.
In this latter case, however, a significant speed-up can be obtained only at pretty low accuracy, whatever the technique used.

\subsection{Structured data}

\begin{figure*}[t]
\centerline{
\includegraphics[width=5.6cm]{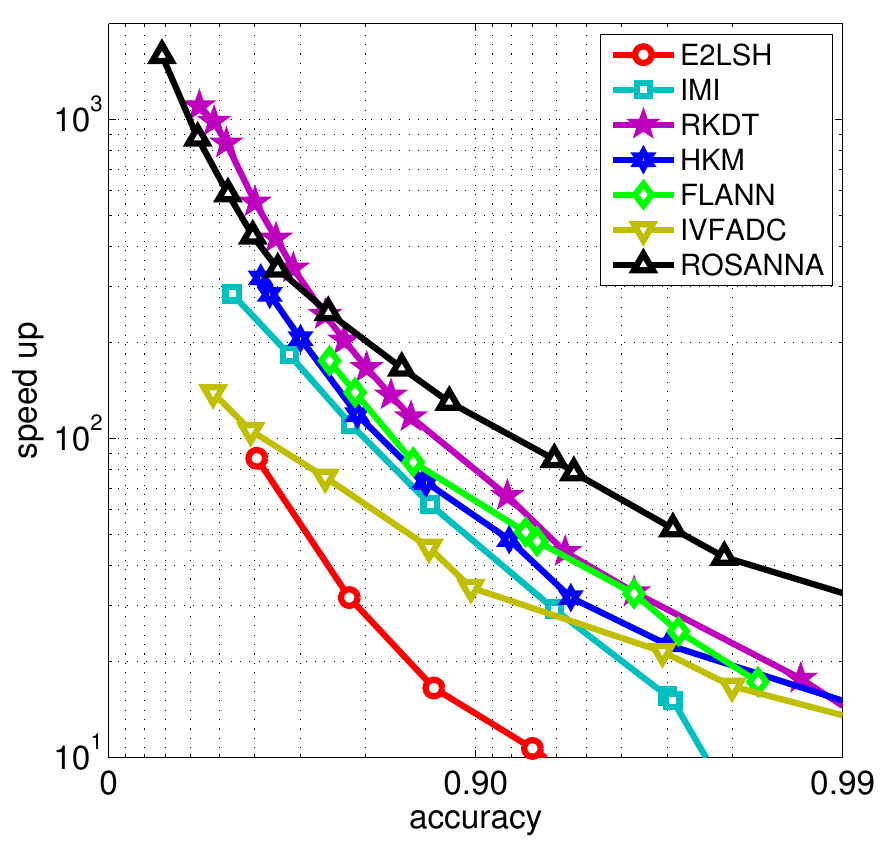}
\includegraphics[width=5.6cm]{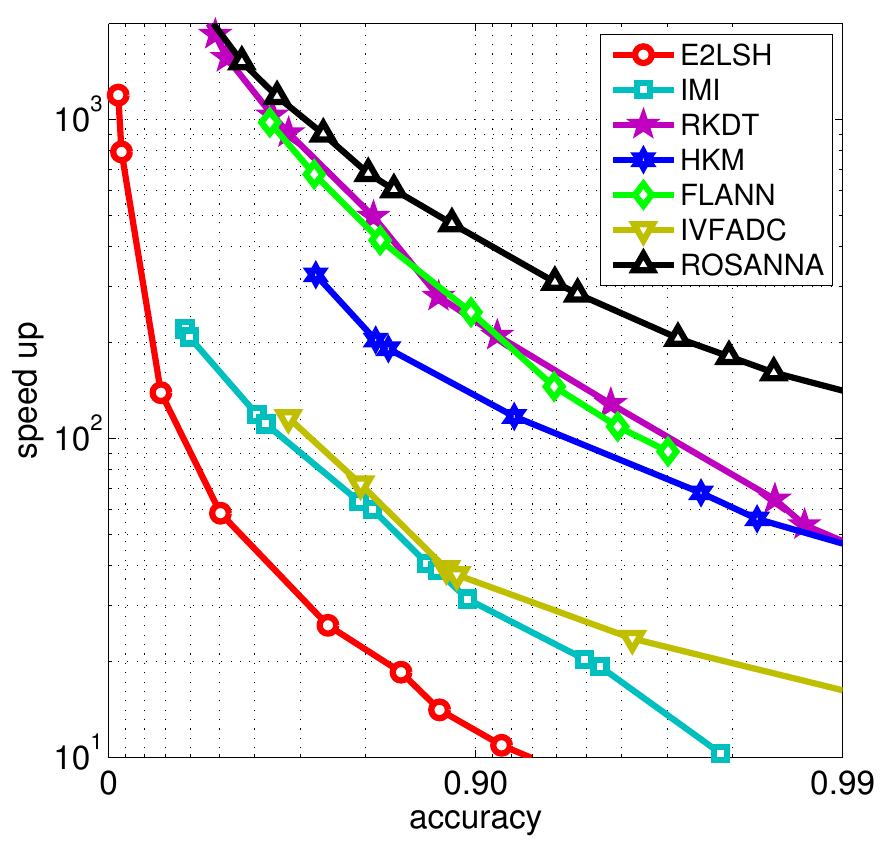}
\includegraphics[width=5.6cm]{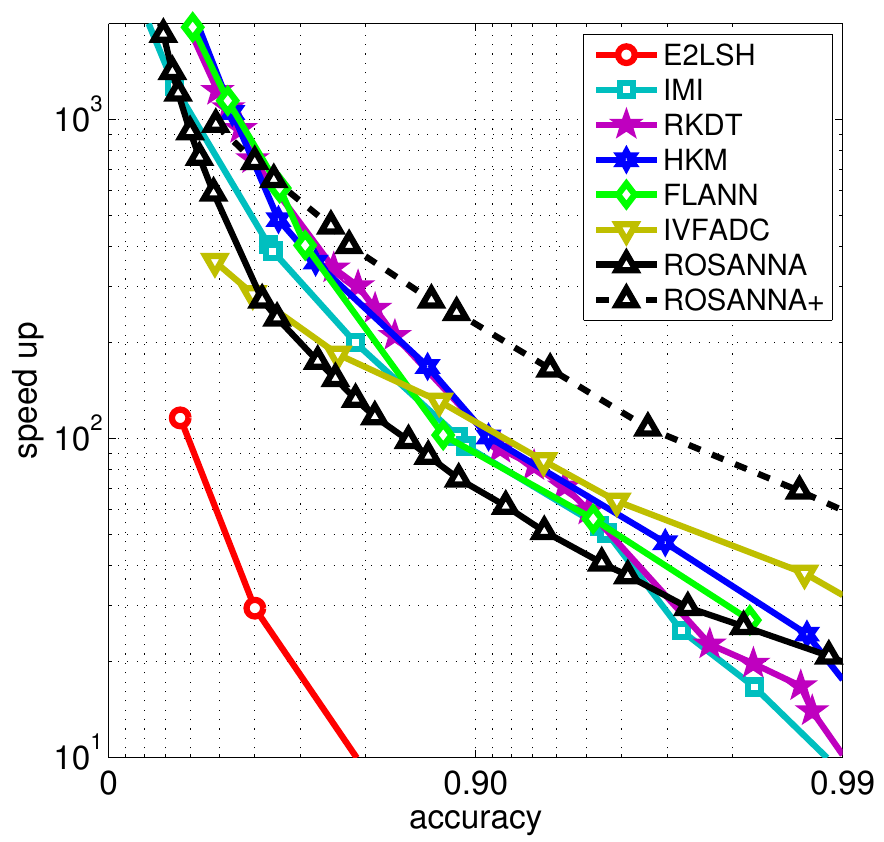}
}


\vspace{1mm}
\caption{Experimental results with real-world data. Left: 100K UBC SIFT; center: 60K MNIST; right: 1M IRISA SIFT.
ROSANNA outperforms almost uniformly all reference techniques in the first two cases. The same happens in the third case after k-means clustering (ROSANNA+).
}
\label{fig:SIFT}
\end{figure*}

Previous experiments confirm that ROSANNA works very well with unstructured data.
It can be argued, however, that most real-world datasets are highly structured, and often have large dimensionality.
Therefore, we now consider some popular structured sources,
SIFT descriptors \cite{Lowe2004}, and MNIST images of handwritten digits, often used to test the performance of ANN search algorithms.
In particular we will use
the 100K-vector UBC SIFT dataset\footnote{\url{http://people.cs.ubc.ca/~mariusm/uploads/FLANN/datasets/sift100K.h5}},
the 1M-vector IRISA SIFT dataset\footnote{\url{http://corpus-texmex.irisa.fr/}},
and the 60K-vector MNIST database\footnote{\url{http://yann.lecun.com/exdb/mnist/}},
with the train/test split coming with each one.
SIFT vectors have length 128, while MNIST images comprise 784 pixels.
In both cases, we search the datasets based on reduced-dimensionality vectors.
First, we compute the PCA and project the points on the new basis,
and then classify the data based only on the first 16 components,
which account for a large fraction of the energy (about 70\% for the SIFT datasets and 60\% for the MNIST database).
In any case,
the NN is searched among the selected candidates by computing distances over all components,
using partial distance elimination (PDE) \cite{ChangDa1985} to speed-up the process.

\begin{figure}[t]
\centerline{
\includegraphics[width=4.2cm]{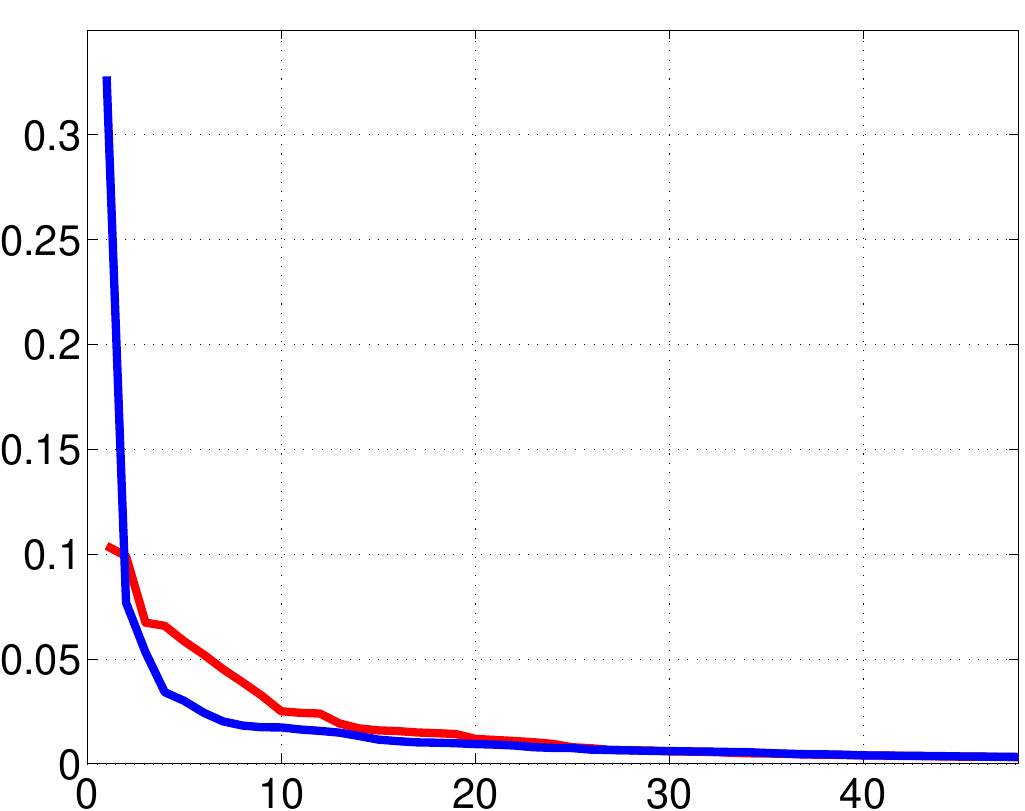}
\includegraphics[width=4.2cm]{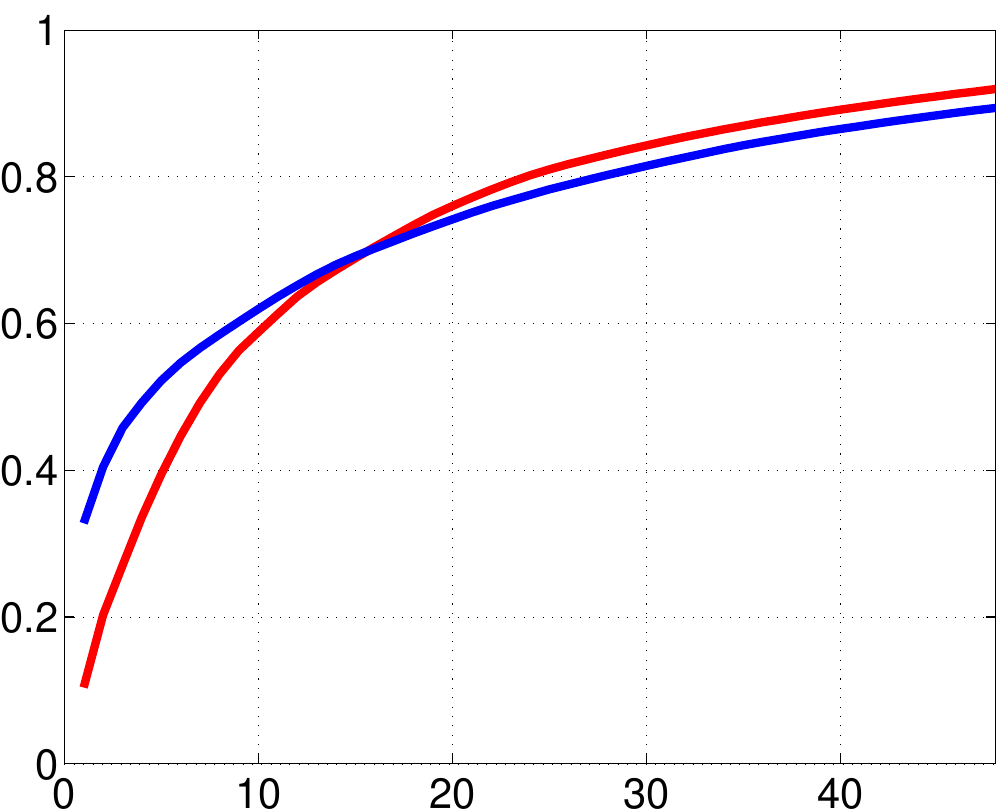}}
\caption{Normalized eigenvalues (left) and their cumulative sum (right) for two SIFT datasets.
Energy distribution is highly skewed, with more than 70\% of the energy in the first 16 components.
The intrinsic data dimensionality is much smaller than 128.
}
\label{fig:SIFT_eigenvalues}
\end{figure}

The use of PCA is motivated by the need to reduce complexity and, {\it a posteriori}, by experimental evidence.
However, it is also justified by the observation that the intrinsic dimensionality of real-world data is typically much smaller than their nominal dimensionality.
SIFT descriptors, for example, have a nominal dimensionality of 128, but the components are also strongly correlated.
In Fig.\ref{fig:SIFT_eigenvalues} we show, for both the UBC and the IRISA datasets,
the first 48 normalized eigenvalues $\lambda_i$ (a), which account for the distribution of energy among the vector components after taking the PCA, and their cumulative sum (b).
In both cases, and especially for IRISA, the distribution is very far from uniform,
with most eigenvalues very close to zero, and 70\% of the total energy in the first 16 components.
Therefore, the intrinsic data dimensionality is much smaller than 128.
A rough estimate is
\begin{equation}
    D^* = 2^{H(p)}
\end{equation}
where $H(p)$ is the informational entropy, and $p_i=\lambda_i/\sum_i(\lambda_i)$.
With this definition, the intrinsic dimensionality drops to 38.69 for UBC and 27.94 for IRISA.
Of course, this estimate neglects {\em non-linear} dependencies, quite significant for SIFT data (and MNIST as well), so the true intrinsic dimensionality is arguably even smaller.

Results are reported in Fig.\ref{fig:SIFT}.
In general, a much larger speed-up is obtained w.r.t. to the case of unstructured data (notice the decade shift on the y-axis)
and no obvious loss of accuracy is observed due to the classification performed only on 16 components.
On both the UBC SIFT and MNIST datasets,
ROSANNA outperforms all reference techniques in the medium-accuracy and especially high-accuracy range.
In the 0.9--0.99 accuracy range, it is about twice as fast as the best competitors.
The situation changes with the IRISA SIFT dataset, where all techniques, including ROSANNA (and except E2LSH), provide a comparable performance.
The reason lies in the much stronger structure of the IRISA data,
where the first PCA component accounts for almost 33\% of the total energy, as opposed to just above 10\% for UBC data
(the dataset size is, instead, immaterial, as the same behavior is observed with 100K vectors).
This is not surprising, since ROSANNA is not designed to exploit data dependencies.
In this case, the best performance is provided by IVFADC,
which performs a preliminary k-means clustering, thus exploiting major data dependencies,
before resorting to product quantization within a restricted number of clusters.
We resorted therefore to a similar solution to adapt ROSANNA to the case of highly dependent data.
Data are clustered off-line by k-means.
At search time, the query is compared with the cluster centroids, and only the nearest clusters are analyzed with ROSANNA,
collecting the candidates that are eventually searched for the NN.
The overall search time is roughly halved w.r.t. the basic version, providing much better results than all references, including IVFADC, especially at high accuracies.
Preliminary k-means clustering is instead ineffective with the less structured UBC SIFT and MNIST data.

We conclude this analysis by showing, in Tab.\ref{tab:parameters}, inspired to Tab.I of \cite{Muja2014},
some numerical performance figures of ROSANNA for the 100K UBC-SIFT dataset as a function of its main parameters.
In the first line we consider a pivot configuration with speed-up 100,
while the following six lines (labeled $G-$, $G+$, $R-$, $R+$, $C-$, $C+$) provide some insight into the effect of increasing/decreasing only one of the parameters at a time w.r.t. the pivot.
By increasing $G$, narrower cones are generated, leading to faster search (remember that the other parameters are fixed) but lower accuracy,
while the opposite happens when $G$ decreases.
Operating on $R$ and $C$ produces similar effects, slower search and higher accuracy when the parameter grows, and the opposite when it decreases.
In all these cases, there is a nearly linear relationship between accuracy and speed-up, when one grows the other decreases.
Memory overhead, instead, exhibits a more varied behavior.
It remains constant when operating on $C$, since the data structure does not change.
It is positively correlated with speed when operating on $G$, because faster search is obtained by defining more cones.
On the contrary,
it is negatively correlated with speed when operating on $R$, because faster search is obtained by using less rotations.
Therefore,
one can keep memory low both when high-speed is required (reducing $R$) and when high accuracy is required (reducing $G$).
This is reflected in the next two configurations,
selected to guarantee high speed, and high accuracy, respectively, where memory overhead is always relatively low.
The last configuration instead has almost no memory overhead and still a good performance.

\renewcommand{\ru}{\rule{0mm}{3.5mm}}
\renewcommand{\tabcolsep}{3.5pt}
\begin{table}[t]
\centering
{\footnotesize
\begin{tabular}{|c|ccc||c|c|c|c|}\hline
\ru \multirow{2}{*}{configuration} & \multicolumn{3}{|c||}{parameters} & \multirow{2}{*}{accuracy} & search   &  memory  & ~build~ \\                 
\ru              & $G$ & $R$ & $C$ &                                                               & speed-up & overhead &   time  \\ \hline \hline   
\ru  pivot       &   4 &   8 &   4 &   0.905 &      100 &     0.36 &   0.36  \\ \hline          
\ru  $G-$        &   3 &   8 &   4 &   0.961 &       37 &     0.16 &   0.35  \\                 
\ru  $G+$        &   5 &   8 &   4 &   0.814 &      168 &     0.69 &   0.36  \\ \hline          
\ru  $R-$        &   4 &   4 &   4 &   0.788 &      180 &     0.18 &   0.26  \\                 
\ru  $R+$        &   4 &  16 &   4 &   0.966 &       54 &     0.71 &   0.57  \\ \hline          
\ru  $C-$        &   4 &   8 &   2 &   0.841 &      145 &     0.36 &   0.35  \\                 
\ru  $C+$        &   4 &   8 &   8 &   0.946 &       66 &     0.36 &   0.36  \\ \hline \hline   
\ru  high speed  &   6 &   2 &  16 &   0.595 &      404 &     0.24 &   0.20  \\ 
\ru  high accu.  &   3 &  16 &   8 &   0.999 &       14 &     0.30 &   0.56  \\ 
\ru  low memory  &   3 &   1 & 128 &   0.901 &       18 &     0.03 &   0.17  \\ \hline          
\end{tabular}
}
\vspace{2mm}
\caption{Performance figures for various parameter configurations.
Memory overhead and build time are relative to dataset occupation and linear search time for the test set, respectively.}
\label{tab:parameters}
\end{table}

\subsection{Fast copy-move forgery detection}

{
With the diffusion of powerful image editing tools,
image manipulation, often with malicious and dangerous aims, has become easy and widespread.
Copy-move is one of the most common attacks, where a piece of the image is cut and paste somewhere else in the same image to hide some undesired objects.
Using material drawn from the same target image, in fact, raises the likelihood to escape the scrutiny of a casual observer.

The most effective copy-move detectors \cite{Christlein2012, Ryu2013, Cozzolino2015} are based on dense feature matching.
A feature is associated with each block, and the most similar feature is searched for over the whole image.
Eventually, a dense field of offsets linking couple of pixels is obtained which, after some suitable post-processing, may reveal the presence of near-duplicate regions.
By using scale/rotation invariant features, copy-moves can be effectively detected even in the presence of geometrical distortions.

Feature matching is the most computation-intensive phase of copy-move detection algorithms.
In \cite{Cozzolino2015}, this task is carried out by resorting to PatchMatch \cite{Barnes2009},
a fast randomized iterative algorithm which outputs a smooth and accurate offset field.
Nonetheless, this process turns out to be relatively slow, since several iterations of PatchMatch are needed for convergence.
This may be a serious problem when a large number of images must be analyzed in search of possible manipulations, especially with today's multi-million pixel images.
To speed-up this phase, one can improve the initialization of PatchMatch (random in the original algorithm) by means of approximate NN search tools, as done for example in \cite{Korman2011}.
This may be effective, indeed, if the ANN algorithm is itself fast and reasonably accurate.
We show next that ROSANNA, with a suitable choice of the parameters, may accomplish very well this task.

We use the copy-move detection algorithm proposed in \cite{Cozzolino2015}, referring the reader to the original paper for all details.
For each 16$\times$16-pixel block we compute the first 16 Zernike moments, which represent the associated rotation-invariant feature.
Based on these features, we then build the offset field by means of PatchMatch, and finally use a suitable post-processing to extract possible near-duplicates.

\begin{figure}[t]
\centerline{
\includegraphics[width=4.4cm]{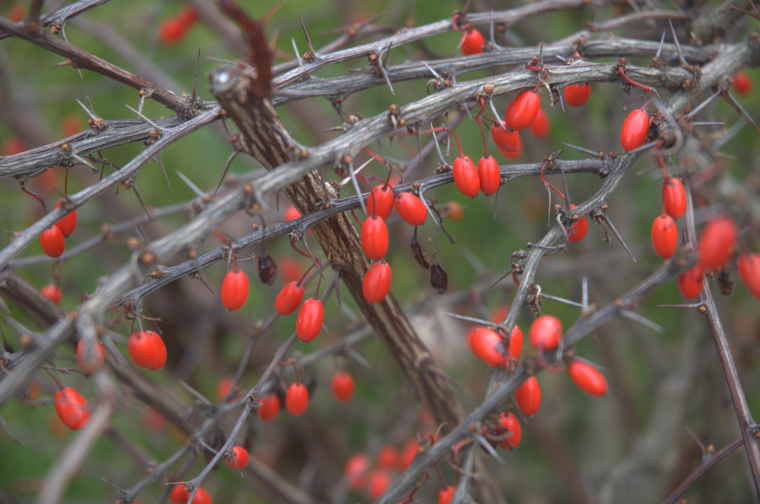}
\includegraphics[width=4.4cm]{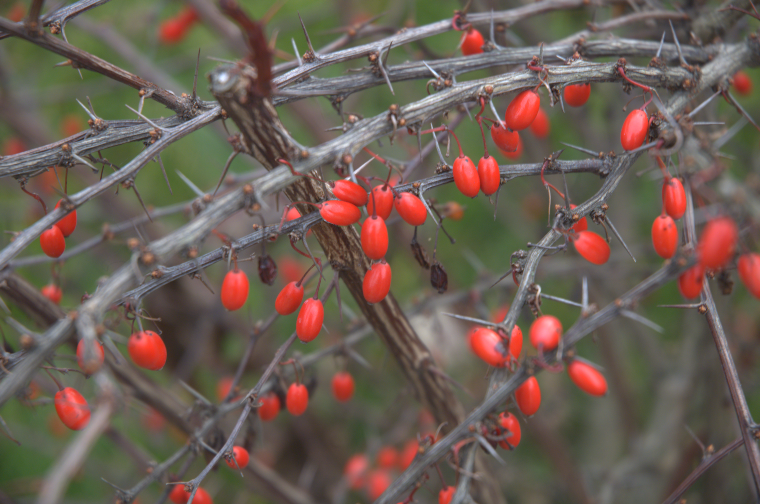}   }

\vspace{1mm}
\centerline{
\includegraphics[width=4.4cm]{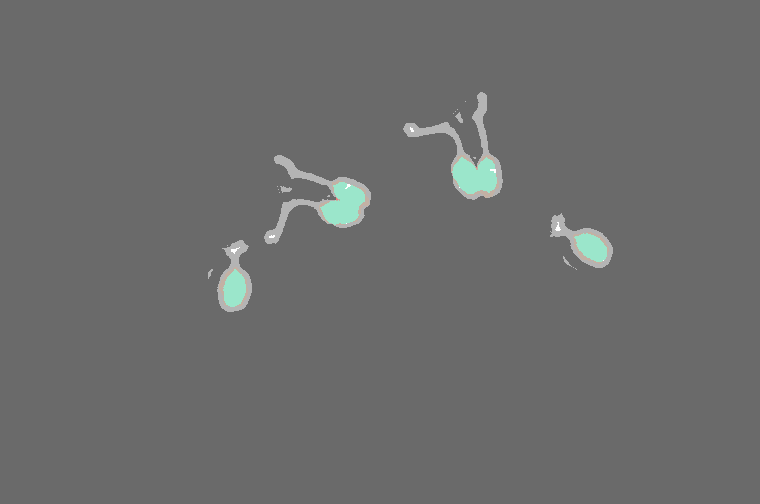}
\includegraphics[width=4.4cm]{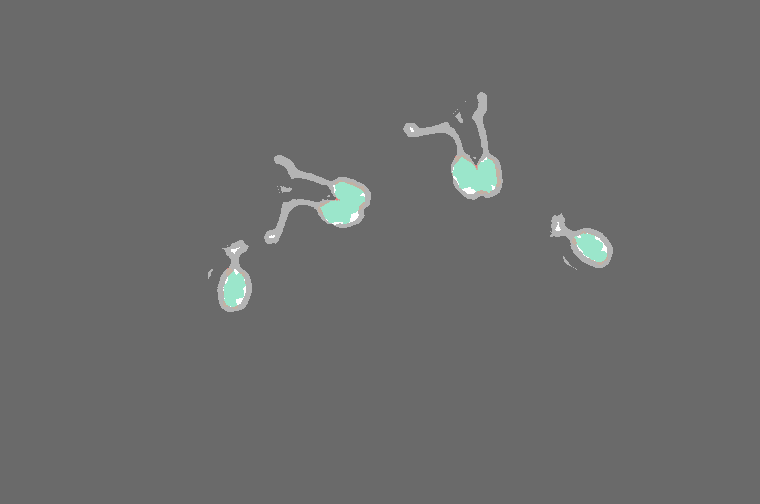}   }
\caption{An example of copy-move forgery detection results.
Top: original and forged images. Bottom: color-coded masks output by the original algorithm (left) and its modified version with ANN initialization (right).
In both cases, despite rotation, all copied regions are detected (green) with very high accuracy, and no false alarm (red) is present. }
\label{fig:CMFD}
\end{figure}

In Tab.\ref{tab:CMFD} we report the average results (CPU-time and accuracy)
observed on a small set of test images when using PatchMatch with random initialization (the original algorithm) and with the initialization provided by ROSANNA.
Images were drawn from the FAU database\footnote{\url{http://www5.cs.fau.de/}} used in \cite{Christlein2012}, and have an average size of 6 Mpixel.
All of them have been manipulated by copy-moving some parts, either to replicate or to hide some objects.
An example is show in Fig.\ref{fig:CMFD}.

With random initialization, eight iterations of PatchMatch are necessary to ensure convergence to an accurate offset field.
As a consequence, the processing time is dominated by the matching phase.
Using ROSANNA to initialize the offset field has some extra costs
due to the need of creating the support data structure and to perform the ANN search.
This latter is reduced through a suitable choice of the parameters, $G=11, R=1, C=1$ and by moderate subsampling $(1:3)\times(1:3)$.
Although only a subset of the offsets are initialized, due to subsampling,
this is more than enough to ensure convergence to a good offset field with only two iterations of PatchMatch.
In addition, the random search phase of PatchMatch is skipped altogether, reducing the cost of a single iteration.
Eventually, the extra cost of initialization is more than compensated by the saving in matching,
leading to an overall 60\% cut in CPU-time for the same accuracy.
Fig.\ref{fig:CMFD} shows in the bottom the output masks provided by the two version of the copy-move detector, which are barely distinguishable.

\begin{table}[t]
\centering
{\footnotesize
\begin{tabular}{|l|rr|rr|}\hline
\ru Task                  & \multicolumn{2}{c|}{Random initialization} & \multicolumn{2}{c|}{ANN initialization} \\ \hline \hline
\ru Feature extraction    &                 6.19~ &          (4.8\%)~~ &              6.19~   &       (11.8\%)~~ \\ \hline
\ru Initialization        &                    \multicolumn{2}{c|}{--} &             27.43~   &       (52.2\%)~~ \\ \hline
\ru PatchMatch            &               100.19~ &         (84.9\%)~~ &              7.66~   &       (14.6\%)~~ \\ \hline
\ru Post-processing       &                13.31~ &         (10.3\%)~~ &             11.22~   &       (21.4\%)~~ \\ \hline
\ru TOTAL CPU-time        &                \multicolumn{2}{c|}{129.69} &              \multicolumn{2}{c|}{52.50} \\ \hline \hline
\ru Average accuracy (FM) &                 \multicolumn{2}{c|}{0.977} &              \multicolumn{2}{c|}{0.971} \\ \hline
\end{tabular}
}
\vspace{2mm}
\caption{Average CPU-time and accuracy for copy-move forgery detection on 6 Mpixel images.}
\label{tab:CMFD}
\end{table}
}


\section{Conclusions and future work}


We developed ROSANNA starting from the analysis of Fig.\ref{fig:scatter_sort_nosort},
and noting that order statistics allow one to induce structure in otherwise unstructured data, so as to speed up ANN search.
Under a different point of view,
ROSANNA is just spherical LSH where, however, multiple alternative partitions of the space of directions are available.
Experiments show ROSANNA to provide a state-of-the-art performance on unstructured data.
Such good results are confirmed also on real-world SIFT and MNIST data.
However, when data are highly structured, some suitable data-dependent preprocessing is needed.
In this work we provided a simple solution to this problem, but there is certainly much room for further improvements.
Finally,
we illustrated ROSANNA's potential to address real-world image processing problems by considering the copy-move forgery detection problem.

Due to its simple conception and implementation,
ROSANNA may represent a precious tools in a wide range of image processing problems where nearest neighbor search is involved.
Research is under way to declinate the same basic concepts in the context of large database image retrieval,
defining reliable proxy distances based on compact codes derived from order-statistics.

\begin{appendices}
\newcommand{\X}{{\bf X}}
\newcommand{\Y}{{\bf Y}}
\newcommand{\norm}[1]{{\lVert #1 \rVert}}

\section{Basic results on order statistics}

Let
\[
    \X = \{X_1, \ldots, X_K\}
\]
be a vector of i.i.d. zero-mean random variables with marginal pdf $f_X(x)$.
The order statistics $X_{(1)}, \ldots, X_{(K)}$ are the random variables obtained by sorting the values of $\X$ in descending\footnote{More often,
ascending order is used, but the two choices are equivalent and we prefer to remain coherent with ROSANNA's functioning.} order.
For example
\[
    X_{(1)} = \max(X_1, \ldots, X_K)
\]
and
\[
    X_{(K)} = \min(X_1, \ldots, X_K)
\]

\noindent
\begin{figure}[t]
\setlength{\unitlength}{1mm}
\begin{picture}(80,30)(+00,+00)
\put(06,24){\vector(1,0){72}}
\put(36,27){\footnotesize $(K \minus i)$}
\put(54,22){\line(0,1){04}}
\put(53,19){\footnotesize $x$}
\put(55,27){\footnotesize $(1)$}
\put(60,22){\line(0,1){04}}
\put(59,19){\footnotesize $x \plus dx$}
\put(66,27){\footnotesize $(i \minus 1)$}
\put(12,24){\circle*{1.2}} \put(15,24){\circle*{1.2}} \put(21,24){\circle*{1.2}} \put(27,24){\circle*{1.2}}
\put(36,24){\circle*{1.2}} \put(39,24){\circle*{1.2}} \put(42,24){\circle*{1.2}} \put(48,24){\circle*{1.2}}
\put(51,24){\circle*{1.2}} \put(57,24){\circle*{1.2}} \put(66,24){\circle*{1.2}} \put(69,24){\circle*{1.2}}

\put(06,06){\vector(1,0){72}}
\put(14,09){\footnotesize $(K \minus j)$}
\put(24,04){\line(0,1){04}}
\put(23,01){\footnotesize $y$}
\put(25,09){\footnotesize $(1)$}
\put(30,04){\line(0,1){04}}
\put(29,01){\footnotesize $y \plus dy$}
\put(36,09){\footnotesize $(j \minus 1 \minus i)$}
\put(54,04){\line(0,1){04}}
\put(53,01){\footnotesize $x$}
\put(55,09){\footnotesize $(1)$}
\put(60,04){\line(0,1){04}}
\put(59,01){\footnotesize $x \plus dx$}
\put(66,09){\footnotesize $(i \minus 1)$}
\put(12,06){\circle*{1.2}} \put(15,06){\circle*{1.2}} \put(21,06){\circle*{1.2}} \put(27,06){\circle*{1.2}}
\put(36,06){\circle*{1.2}} \put(39,06){\circle*{1.2}} \put(42,06){\circle*{1.2}} \put(48,06){\circle*{1.2}}
\put(51,06){\circle*{1.2}} \put(57,06){\circle*{1.2}} \put(66,06){\circle*{1.2}} \put(69,06){\circle*{1.2}}
\end{picture}
\caption{Reference geometry for computing order statistics of the first (top) and second (bottom) order. In the examples $K=12, i=3, j=9$.}
\label{fig:OS_geometry}
\end{figure}
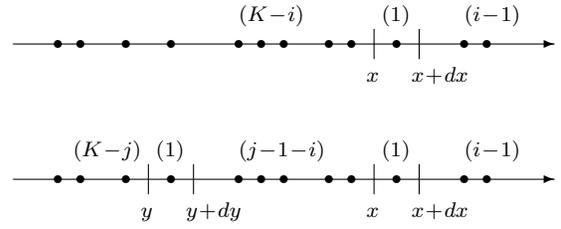

To obtain the marginal pdf of the $i$-th order statistics let us compute the probability
\[
    \Pr(X_{(i)} \in [x,x+dx]) = f_{X_{(i)}}(x) dx
\]
where $dx \to 0$.
For this event to happen,
$i-1$ components of vector $X$ must be larger than $x+dx$,
$K-i$ of them must be smaller than $x$,
and exactly one of them must fall in the interval $[x,x+dx]$ (see Fig.\ref{fig:OS_geometry} top).
Since the components are i.i.d.,
by taking into account the different combinations through a multinomial coefficient, and neglecting $O(dx^2)$ terms, it results
\begin{IEEEeqnarray}{rCl}
    \label{eq:OS_pdf_1}
    f_{X_{(i)}}(x) & =      & \frac{K!}{(K-i)!(i-1)!} \\[1mm]
                   & \times & [F_X(x)]^{K-i} [1-F_X(x)]^{i-1} f_X(x) \nonumber
\end{IEEEeqnarray}
All joint statistics can be computed in the same way.
In particular, for the second-order joint pdf, taking $i<j$, it results (see Fig.\ref{fig:OS_geometry} bottom)
\begin{IEEEeqnarray}{rCl}
    \label{eq:OS_pdf_2}
    f_{X_{(i)}X_{(j)}}(x,y) & =      & \frac{K!}{(K-j)!(j-1-i)!(i-1)!} \\[1mm]
                            & \times & [F_X(y)]^{K-j} [F_X(x)-F_X(y)]^{j-1-i} \nonumber \\[1mm]
                            & \times & [1-F_X(x)]^{i-1} f_X(x)f_X(y) u(x \minus y) \nonumber
\end{IEEEeqnarray}

\section{Theoretical bounds}
We want to characterize ROSANNA in terms of its collision probability for unstructured data.

Let
\[
    \X = \{X_1, \ldots, X_K\}
\]
be a vector of i.i.d. zero-mean symmetric random variables with marginal pdf $f_X(x)$.
Then, let
\[
    \Y = \{Y_1, \ldots, Y_K\}
\]
be a second vector which, given $\X = \x$, is uniformly distributed on the hypersphere of radius $r$ centered on $\x$, that is
\[
    f_{\Y}(\y|\x) = \frac{1}{S^K(r)} \delta(\norm{\y-\x}-r)
\]
where $S^K(r)$ is the measure of the $K$-d hypersphere of radius $r$, and $\delta(\cdot)$ is the Dirac delta function.
We want to compute, for any radius $r$, the probability that $\X$ and $\Y$ belong to the same cone
\[
    p_{\rm coll}(r) = \Pr[h(\X)=h(\Y)]
\]
where the hash function $h(\cdot)$ associates a vector to a given cone based on the index and sign of its $G$ largest components.
Rather than the probability of collision, we will consider its complement
\[
    p_{\rm cross}(r)=1-p_{\rm coll}(r)
\]
which is the probability that point $\Y$ lies across any of the boundaries of $\X$'s cone.

\subsection{The 2d case}

\begin{figure}[t]
\setlength{\unitlength}{1mm}
\begin{picture}(60,45)(-15,+15)

\put(10,30){\line(1,0){18}}
\put(35,30){\vector(1,0){15}}
\put(50,28){\footnotesize $y_1$}
\put(20,20){\vector(0,1){30}}
\put(21,51){\footnotesize $y_2$}
\put(30,29){\footnotesize $c_1$}
\put(48,35){\footnotesize $\Omega$}
\put(35,20){\footnotesize $c_1 \!\setminus\! \Omega$}
{\thicklines
\put(20,30){\line(+1,+1){25}}
\put(20,30){\line(+1,-1){15}}
\put(20,30){\line(-1,-1){10}}
\put(20,30){\line(-1,+1){10}}
}
\put(40,42){\circle{14}}
\put(40,42){\circle*{1}}
\put(41,40){\footnotesize $\x$}
\put(40,42){\line(-1,+1){4}}
\put(39,44){\footnotesize $\Delta$}
\put(40,42){\line(-6,1){7}}
\put(36,40){\footnotesize $r$}
\put(35,43){\footnotesize $\theta$}
\end{picture}
\caption{Geometry of the problem in 2d. }
\label{fig:2d_example}
\end{figure}
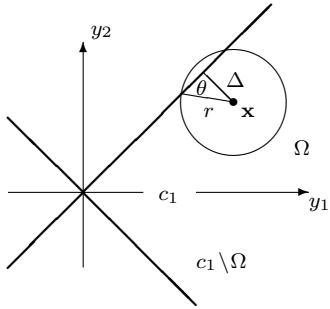

Let us begin by considering the simplest non-trivial case of $K=2$ and $G=1$.
Fig.\ref{fig:2d_example} provides a pictorial description of our problem in this setting.
In the example, point $\x$ belongs to cone $c_1$ where the first component is the largest, and it is positive.
In addition, we focus on the half-cone $\Omega$ where also the second component of $\x$ is positive
\[
    \Omega = \{\x \in \RR^2: x_1 \geq x_2 \geq 0\}
\]
However, thanks to symmetry, all following arguments apply equally well to all other cones and half-cones with obvious modifications.
Therefore
\begin{equation}
    p_{\rm cross}(r) = p_{\rm cross}(r \given \Omega) = \int_{\Omega} f_{\X|\Omega}(\x) p_{\rm cross}(r \given \x) d\x
    \label{eq:pcross_given_Omega}
\end{equation}
Cone $c_1$ has two boundaries, the lines with equations $y_2=y_1$ and $y_2=-y_1$.
For each $\x$ in $c_1$ let us label the boundaries in order of increasing distance from $\x$, so that
\[
    p_{\rm cross,1}(r \given \x)
\]
is the probability of crossing boundary number 1, the nearest one.
Let $\Delta$ be the distance of $\x$ from this boundary.
Since we restrict attention to $\Omega$, this is
\[
    \Delta=(x_1-x_2)/\sqrt{2}
\]
while more in general it holds
\[
    \Delta=\frac{\max(|x_1|,|x_2|)-\min(|x_1|,|x_2|)}{\sqrt{2}}
\]
If $\Delta<r$, part of the circumference of radius $r$ centered on $\x$ will cross the nearest boundary.
For our hypothesis of uniform distribution of $Y|\x$, the fraction of the circumference past the boundary represents the probability of crossing it
\begin{equation}
    p_{\rm cross,1}(r \given \x) =
    \left\{ \begin{array}{ll}
        \theta(r,\Delta)/\pi & \Delta < r \\
        0                    & \mbox{otherwise}
    \end{array} \right.
    \label{eq:lowerbound_2d}
\end{equation}
where
\begin{equation}
    \theta(r,\Delta) = \arccos\left(\frac{\Delta}{r}\right)
    \label{eq:theta}
\end{equation}
Of course, other points on the circumference may cross the second boundary, and some may cross both.
Therefore (\ref{eq:lowerbound_2d}) is only a lower bound to $p_{\rm cross}(r | \x)$.
A good upper bound is obtained by the sum $p_{\rm cross,1}(r | \x) + p_{\rm cross,2}(r | \x)$, which may be difficult to compute.
Simpler but looser bounds are $2\,p_{\rm cross,1}(r | \x)$ and $1-1/N_C$, with $N_C$ the total number of cones.
In summary
\begin{IEEEeqnarray}{rl}
    p_{\rm cross,1}(r | \x) & \leq p_{\rm cross}(r | \x) \\
                            & \leq \min \left( 2\,p_{\rm cross,1}(r | \x),\frac{N_C \minus 1}{N_C} \right) \nonumber
\end{IEEEeqnarray}
and by averaging over $\X \in \Omega$ we obtain lower and upper bounds for $p_{\rm cross}(r)$.

\begin{figure}[t]
\centerline{
\includegraphics[width=4.2cm]{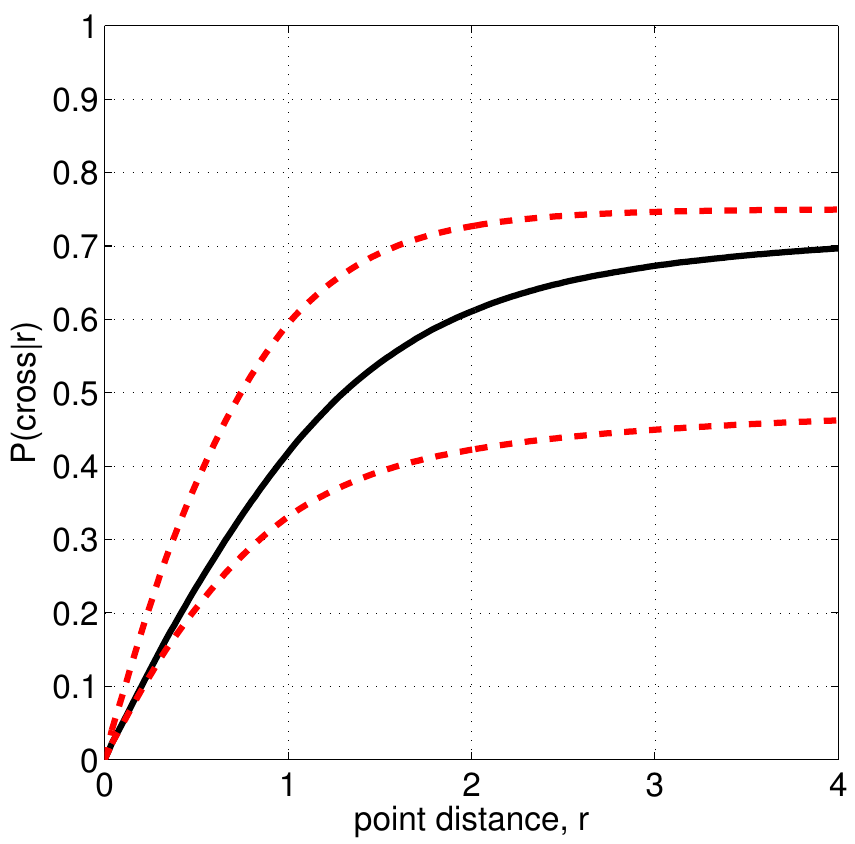}
\includegraphics[width=4.2cm]{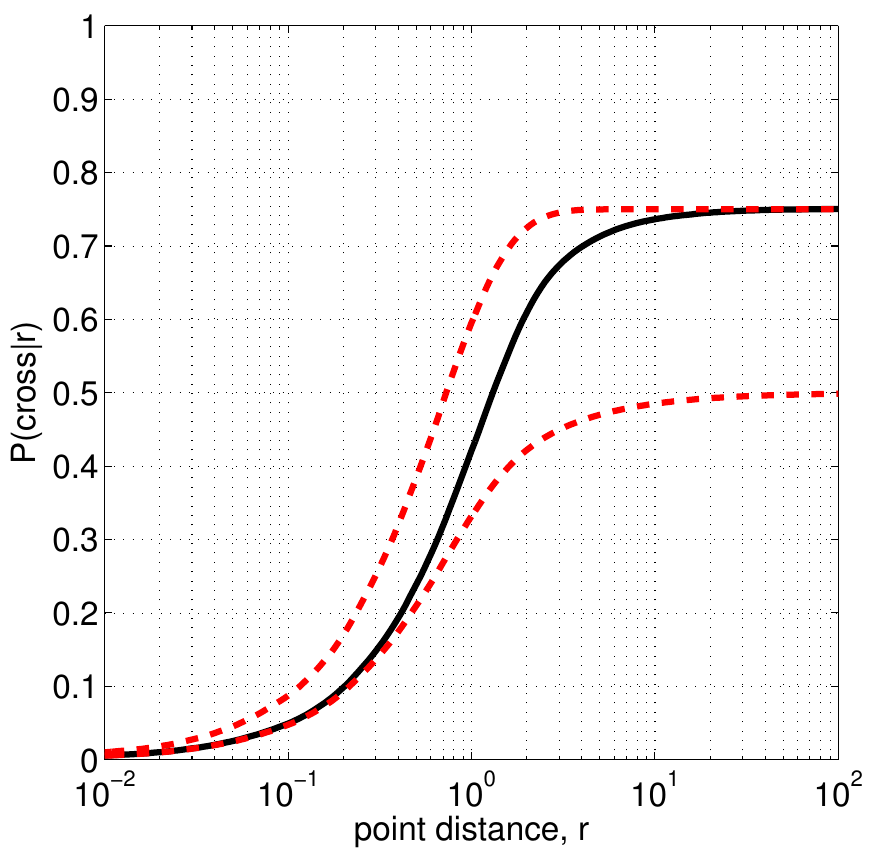}
}
\caption{Theoretical bounds and MonteCarlo estimate for the crossing probability in the case of i.i.d. standard Gaussian RV's, with $K$=2, $G$=1.}
\label{fig:bounds_2d}
\end{figure}

In Fig.\ref{fig:bounds_2d} we plot these upper and lower bounds,
together with the actual crossing probability estimated through MonteCarlo simulation, when the $X_i$'s are standard Gaussian RV's.
As expected,
for small values of $r$ the MonteCarlo estimate is very close to the theoretical lower bound (see also the log-scale plot on the right),
while the gap grows when the distance between the two points, $X$ and $Y$, becomes comparable with their own norm.
When $r \to \infty$, of course, $Y$ belongs to any cone with the same probability,
and the curve approaches the theoretical upper bound of 3/4.

\subsection{The general case}

We now consider the general high-dimensional case, proceeding in the very same way as for the 2d case,
except for some suitable modifications.
The most important difference with respect to the previous case
is that we will resort to order statistics to reduce the final $K$-dimensional integral to a numerically tractable 2d integral.

In this case we have $N_C=\binom{K}{G}2^G$ cones, statistically indistinguishable from one another.
Let us consider the cone $c_1$ where the first $G$ components are also the largest, and they are all positive.
Furthermore, let us restrict attention to the subregion of $c_1$, call it again $\Omega$,
where also the smallest $K-G$ components are all positive
\begin{IEEEeqnarray}{ll}
    \Omega = \{ \x \in \RR^K: & \min(x_1,\ldots,x_G) \geq \max(x_{G+1},\ldots,x_K), \nonumber \\[1mm]
                              & x_i \geq 0, \ls i=1,\ldots,K \}
\end{IEEEeqnarray}
Again, because of symmetry, this restriction is immaterial, and (\ref{eq:pcross_given_Omega}) still holds.
So, we will provide lower and upper bounds for $p_{\rm cross}(r | \x)$, and then integrate over $\Omega$.

Cone $c_1$ is now delimited by $2G(K-G)$ hyperplanes, the hyper-planes with equations $y_i=y_j$ and $y_i=-y_j$, for all $i=1,\ldots,G$ and $j=G+1,\ldots,K$,
and the point $Y$ at distance $r$ from $\x$ leaves the cone only if it crosses at least one of such boundaries.

Let again $\Delta$ be the distance of $\x$ from the closest boundary.
If $\Delta>r$ the probability of crossing that boundary (or any other) is 0.
Otherwise, it can be computed as the ratio between the measure $S^{K,\Delta}(r)$ of the hyper-spherical cap intercepted by a hyperplane at distance $\Delta$ from the center,
and the measure $S^{K}(r)$ of the whole hyper-sphere.
It is known that
\[
    S^K(r) = \frac{2\pi^{K/2}}{\Gamma(\frac{K}{2})}r^{K-1}
\]
with $\Gamma(\cdot)$ the Gamma function,
while the measure of the cap can be computed by integrating over $\theta$ the $(K \minus 1)$-dimensional hyper-spheres of radius $r\sin(\theta)$
\[
    S^{K,\Delta}(r) = \int_0^{\theta(r,\Delta)} S^{K-1}(r\sin(\theta))\,r\,d\theta
\]
where $\theta(r,\Delta)$ is still given by (\ref{eq:theta}).
As for $\Delta$, note that the closest boundary to $\x$ is the hyperplane of equation $y_m=y_M$
where $m$ is the index of the smallest component among the $G$ largest, and $M$ is the index of the largest component among the $K-G$ smallest.
Consequently
\[
    \Delta = \frac{x_m-x_M}{\sqrt{2}}
\]
In summary it results
\[
    p_{\rm cross,1}(r \given \x) =
    \left\{ \begin{array}{ll}
        S^{K,\Delta}(r)/S^{K}(r) & \Delta < r \\
        0                        & \mbox{otherwise}
    \end{array} \right.
\]
Again, $p_{\rm cross,1}(r | \x)$ is a lower bound for $p_{\rm cross}(r | \x)$.
An upper bound can be easily obtained as
\begin{IEEEeqnarray}{rCl}
    p_{\rm cross}(r \given \x) & \leq & \sum_{i=1}^{2G(K-G)} p_{\rm cross,i}(r | \x) \nonumber \\[1mm]
                               & \leq & \min \left( 2G(K-G)p_{\rm cross,1}(r | \x),\frac{N_C-1}{N_C} \right)       \nonumber
\end{IEEEeqnarray}

As the last step of our development, we must compute the integral (\ref{eq:pcross_given_Omega}) over $\X \in \Omega$
which, for $K \gg 1$, is computationally intractable.
However, we are not really interested in the $K$-dimensional integral,
since $p_{\rm cross,1}(r \given \x)$ depends only on the $G$-th and $(G+1)$-st largest components of $\x$ through $\Delta$,
that is,
\begin{IEEEeqnarray}{ll}
    \int_{\Omega} f_{\X|\Omega}(\x) & p_{\rm cross,1}(r \given \x) d\x = \nonumber \\[1mm]
                                    & \int_0^\infty \!\! \int_0^\alpha f_{X_{(G)},X_{(G+1)}|\Omega}(\alpha,\beta) p_{\rm cross,1}(r \given \alpha,\beta) d\beta d\alpha \nonumber
\end{IEEEeqnarray}
where the $X_{(i)}$'s are the order statistics obtained by sorting $\X$ in descending order
(remember that in $\Omega$ this coincides with sorting the vector for descending absolute values).
Therefore we only need the joint pdf of $X_{(G)}$ and $X_{(G+1)}$, which is given by (\ref{eq:OS_pdf_2}).

\begin{figure}[t]
\centerline{
\includegraphics[width=4.2cm]{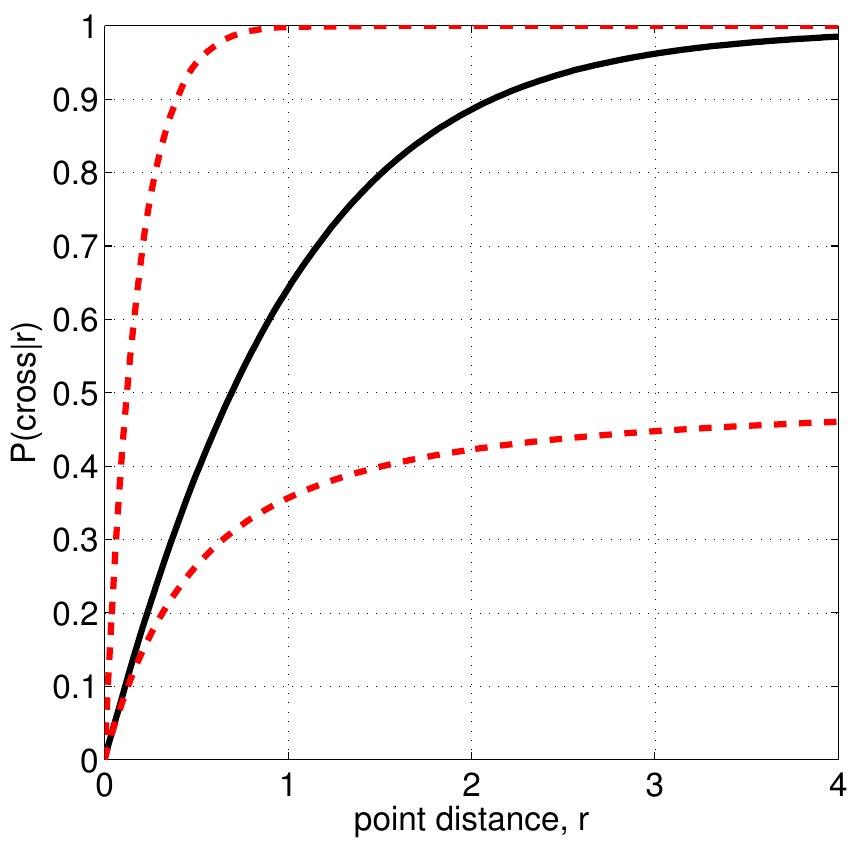}
\includegraphics[width=4.2cm]{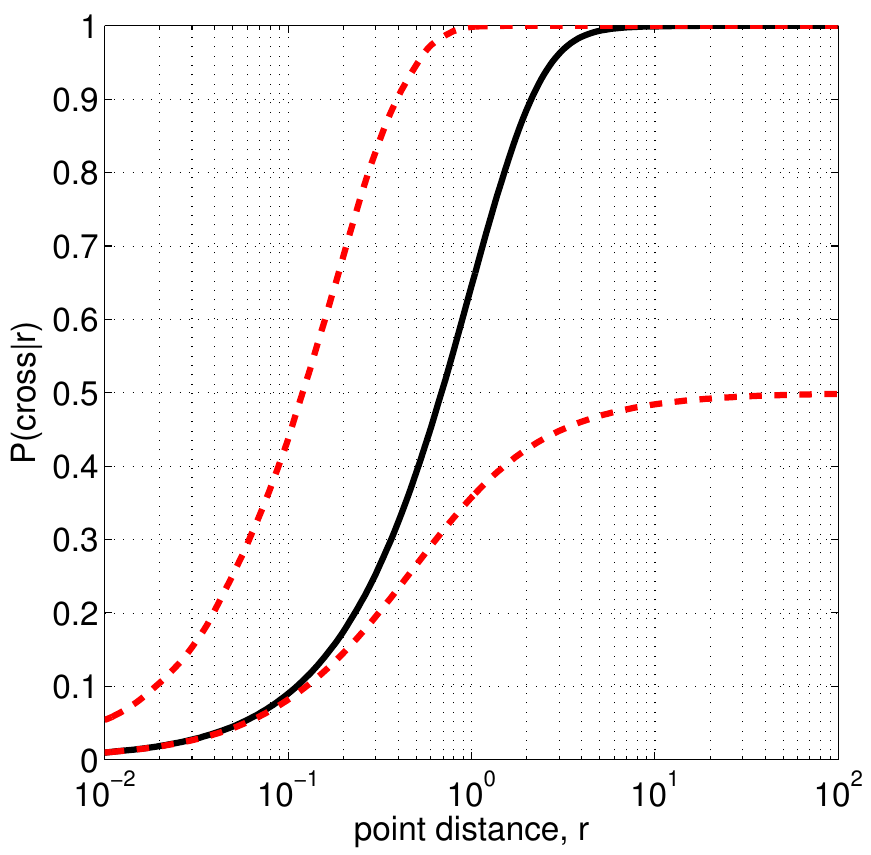}
}
\caption{Theoretical bounds and MonteCarlo estimate for the crossing probability in the case of i.i.d. standard Gaussian RV's, with $K$=16, $G$=4.}
\label{fig:bounds_Kd}
\end{figure}

In Fig.\ref{fig:bounds_Kd}, we plot these upper and lower bounds,
together with the actual crossing probability estimated through MonteCarlo simulation, when the $X_i$'s are standard Gaussian RV's,
for a single high-dimensional case, with $K$=16 and $G$=4.
Again, for small values of $r$ the MonteCarlo estimate is very close to the theoretical lower bound.
The upper bound, instead, is too loose to be of practical guide.

These theoretical results confirm the correctness of the proposed algorithm.
Close points tend to be hashed in the same cell, and the collision probability approaches 1 as the distance goes to 0.
In addition, they can be used to guide the choice of the algorithm parameters.
One could compute upper and lower bounds for each value of $K$ and $G$, and choose the combination of parameters that better meets the problem requirements.
It is worth reminding that these results hold rigorously only for the Gaussian i.i.d. case,
and have been obtained with reference to a simple version of the algorithm, with a single basis and no multi-probe.
Nonetheless, they represent a conceptual support to practical design.


%
%

\end{appendices}

\balance

\bibliographystyle{IEEEbib}
\bibliography{ANN}

\begin{IEEEbiography}[{\includegraphics[width=2in,height=3in,trim = 60mm 0mm 0mm 08mm,clip,keepaspectratio]{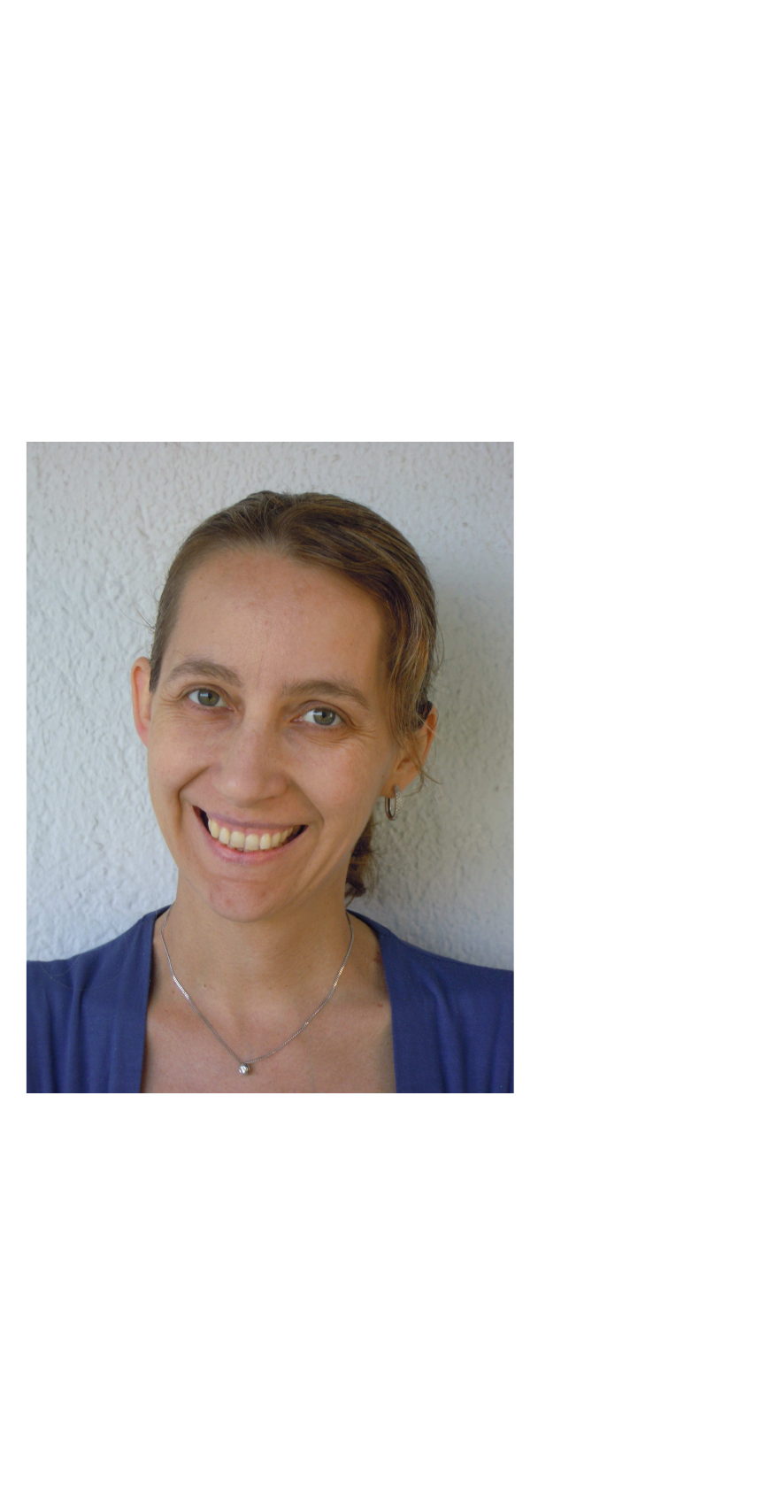}}]
{Luisa Verdoliva}
is Assistant Professor of Telecommunications
with the Department of Electrical Engineering and Information Technology of the University University of Naples Federico II, Naples, Italy.
Her research is on image processing, in particular compression and restoration of remote-sensing images, both optical and SAR, and digital forensics.
Dr. Verdoliva has been elected member of the IEEE Information Forensics and Security Technical Committee for the 2016-2018 period.
\end{IEEEbiography}

\begin{IEEEbiography}[{\includegraphics[width=2in,height=3in,trim = 60mm 0mm 0mm 10mm,clip,keepaspectratio]{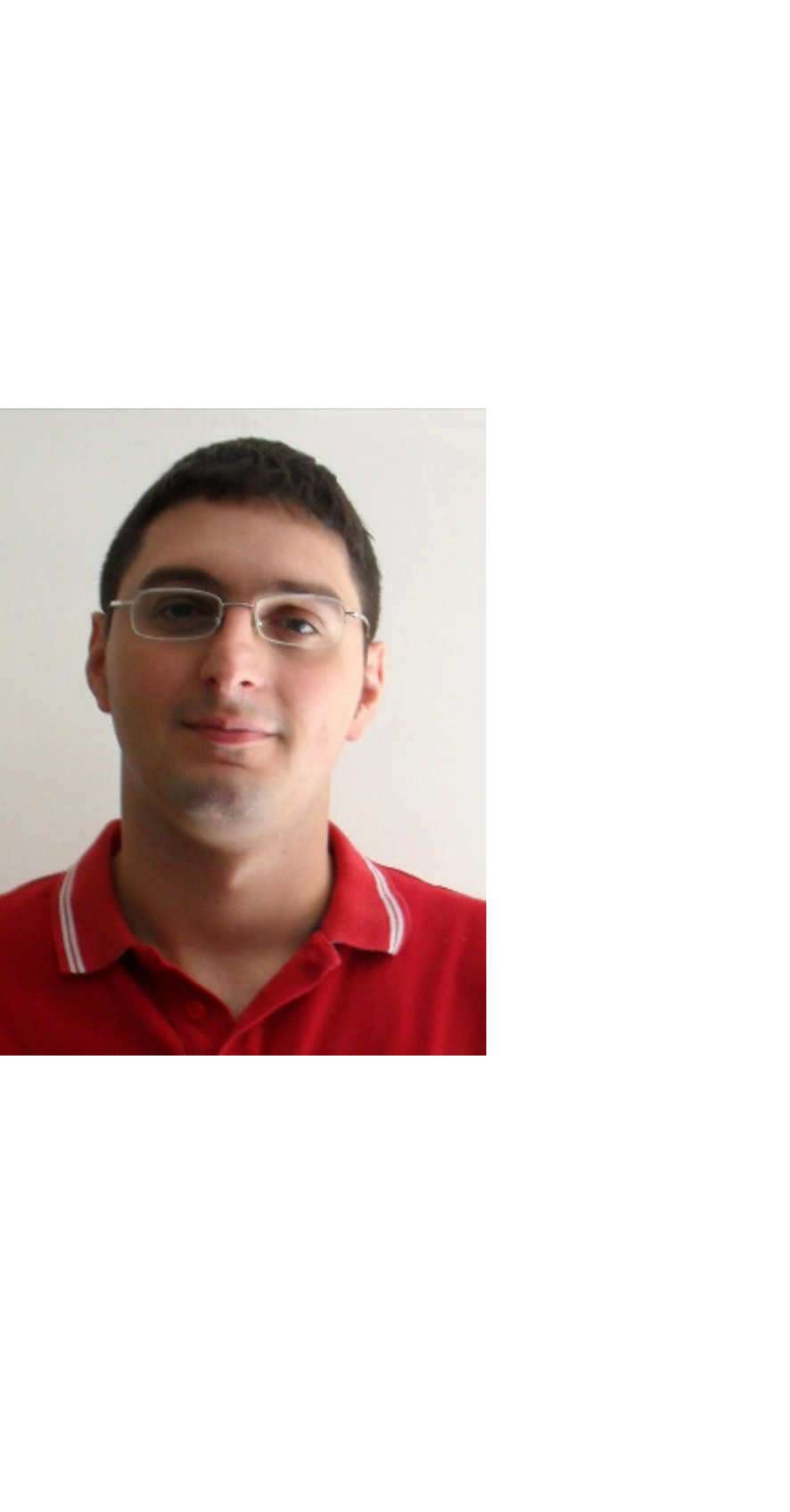}}]
{Davide Cozzolino}
received the Laurea degree in computer engineering and the Ph.D. degree
in information engineering from the
University Federico II of Naples, Italy, in December 2011 and April 2015, respectively.
He is currently a Post Doc at the same University.
His study and research interests include image processing, particularly
forgery detection and localization.
\end{IEEEbiography}

\begin{IEEEbiography}[{\includegraphics[width=2in,height=2.5in,trim = 60mm 0mm 0mm 15mm,clip,keepaspectratio]{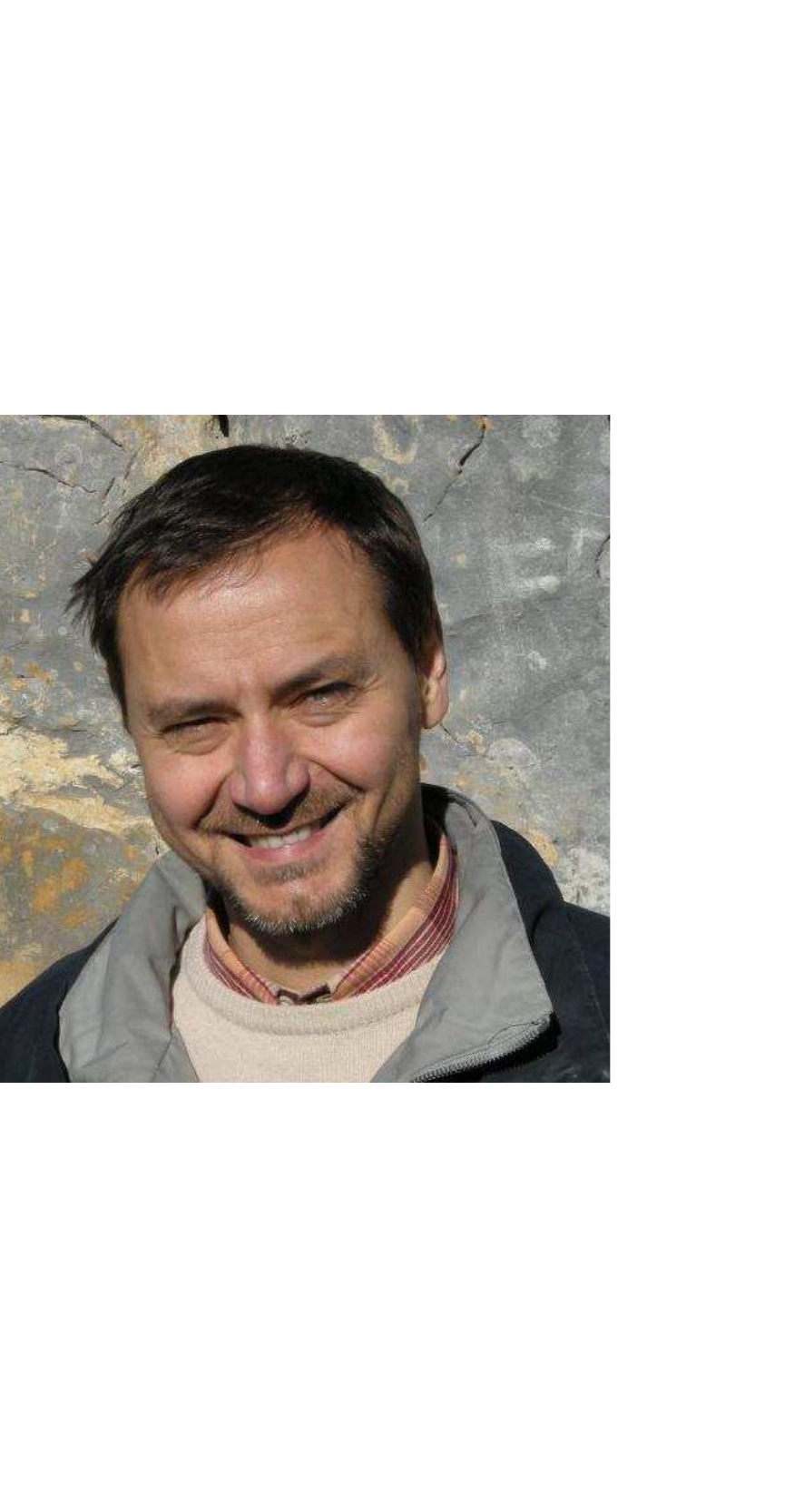}}]
{Giovanni Poggi}
is Full Professor of Telecommunications and vice Director of the
Department of Electrical Engineering and Information Technology of the University University of Naples Federico II, Naples, Italy.
His research interests are in statistical image processing, including compression, restoration, segmentation, and classification,
with application to remote-sensing, both optical and SAR, digital forensics, and biometry.
Prof. Poggi has been an Associate Editor for the IEEE Transactions on Image Processing and Elsevier Signal Processing
\end{IEEEbiography}

\end{document}